\documentclass[journal]{IEEEtran}

\usepackage{multirow}
\usepackage[table,xcdraw]{xcolor}

\usepackage{graphicx}
\usepackage{amsmath}
\usepackage{amsfonts}
\usepackage{algorithm}
\usepackage{amsthm,amssymb}
\usepackage{mathrsfs}
\usepackage{algorithmic}
\usepackage{makecell}
\usepackage{easyReview}
\usepackage{booktabs}
\setcellgapes{3pt}

\usepackage{hyperref}
\hypersetup{hidelinks,
	colorlinks=true,
	allcolors=blue,
	pdfstartview=Fit,
	breaklinks=true}
\setcellgapes{3pt}

\usepackage[justification=centering]{caption}
\usepackage{color}
\usepackage[subfigure]{tocloft}
\usepackage{subfigure}
\usepackage{tabularx}
\usepackage{amsmath, bm}

\usepackage{cite}

\ifCLASSINFOpdf
\else
\fi
\begin{document}
	
	%
	\title{Personalized Federated Learning for GAI-Assisted Semantic Communications}
	
	\author{Yubo Peng, Feibo Jiang, \textit{Senior Member, IEEE}, Li Dong, Kezhi Wang, \textit{Senior Member, IEEE}, and Kun Yang, \textit{Fellow, IEEE}.
	\thanks{
	This work was supported in part by the
	National Natural Science Foundation of China under Grant 41904127 and 62132004, in part by the Hunan Provincial Natural Science Foundation of China under Grant 2024JJ5270, in part by the Open Project of Xiangjiang Laboratory under Grant 22XJ03011, and in part by the Scientific Research Fund of Hunan Provincial Education Department under Grant 22B0663.
	
	Yubo Peng (pengyubo@hunnu.edu.cn) and Kun Yang (kyang@ieee.org) are with the School of Intelligent Software and Engineering, Nanjing University, Suzhou, China.
			
	Feibo Jiang (jiangfb@hunnu.edu.cn) is with School of Information Science and Engineering, Hunan Normal University, Changsha, China. 
	
	Li Dong (Dlj2017@hunnu.edu.cn) is with Changsha Social Laboratory of Artificial Intelligence, Hunan University of Technology and Business, Changsha, China.
	
	Kezhi Wang (Kezhi.Wang@brunel.ac.uk) is with the Department of Computer Science, Brunel University London, UK.
	}

	}

\markboth{Submitted for Review}%
{Shell \MakeLowercase{\textit{et al.}}: Bare Demo of IEEEtran.cls for IEEE Journals}
%



\maketitle

\begin{abstract}
	Semantic Communication (SC) focuses on transmitting only the semantic information rather than the raw data. This approach offers an efficient solution to the issue of spectrum resource utilization caused by the various intelligent applications on Mobile Users (MUs).
	Generative Artificial Intelligence (GAI) models have recently exhibited remarkable content generation and signal processing capabilities, presenting new opportunities for enhancing SC.
	Therefore, we propose a GAI-assisted SC (GSC) model deployed between MUs and the Base Station (BS).
	Then, to train the GSC model using the local data of MUs while ensuring privacy and accommodating heterogeneous requirements of MUs, we introduce Personalized Semantic Federated Learning (PSFL). This approach incorporates a novel Personalized Local Distillation (PLD) and Adaptive Global Pruning (AGP).
	In PLD, each MU selects a personalized GSC model as a mentor tailored to its local resources and a unified Convolutional Neural Networks (CNN)-based SC (CSC) model as a student. This mentor model is then distilled into the student model for global aggregation.
	In AGP, we perform network pruning on the aggregated global model according to real-time communication environments, reducing communication energy.
	Finally, numerical results demonstrate the feasibility and efficiency of the proposed PSFL scheme.
\end{abstract}

\begin{IEEEkeywords}
	Semantic communication; Generative Artificial Intelligence; Federated learning; Internet of Things. 
\end{IEEEkeywords}

\section{Introduction}
As an innovative communication paradigm in 6G, Semantic Communication (SC) becomes one of the intelligent solutions for the spectrum sacrifice caused by the various emerging applications on Mobile Users (MUs) \cite{jiang2024large}.
Different from traditional communication, SC aims to transmit only semantically related information for the respective task/goal \cite{qin2019deep}. 
For example, in the fault detection scenario, the MUs first extract semantic information from the surveillance video by the deployed SC encoder (i.e., semantic and channel encoders), then just transmit slightly semantic information to the Edge Server (ES) deployed on the Base Station (BS). Finally, the received semantic information is decoded by the SC decoder (i.e., semantic and channel decoders) deployed on the ES. Since the transferred data is greatly reduced in SC, the consumption of spectrum resources is also correspondingly greatly decreased. 

The performance of SC is highly dependent on the construction of high-quality SC models, hence many researchers construct SC models based on the  Deep Learning (DL) models. For instance, Xie et al. \cite{xie2021deep} proposed a DL-based SC system, aiming to maximize system capacity and minimize semantic errors in text transmission by restoring the meaning of sentences.
Wang et al. \cite{10001359} optimized DL-based joint source-channel coding by introducing adversarial loss, which better preserved the global semantic information and local texture details of images.
Han et al. \cite{9953316} proposed a novel end-to-end DL-based speech-oriented SC system, utilizing a soft alignment module and a redundancy removal module to extract text-related semantic features while discarding semantically redundant content.
Most of the above works are based on traditional discriminative Artificial Intelligence (AI) methods, which typically involve small models trained for specific application scenarios. This approach inherently limits their adaptability across different environments. Moreover, discriminative AI primarily focuses on learning local and short-term features, leading to challenges such as getting trapped in local minima and exhibiting limited generative capabilities \cite{10638533}.

Generative AI (GAI), as a recent advancement in AI technology, not only possesses remarkable generative capabilities but also exhibits more powerful data processing abilities than discriminative AI. The latest GAI models, such as GPT-4 and LLaMA 3.1 \cite{jiang2023large1}, have been widely applied across various domains. 
Therefore, it has become a recent research topic that constructing SC models based on GAI. 
Du et al. \cite{10158526} designed an AI-generated incentive mechanism based on the diffusion model in full-duplex end-to-end SC to promote semantic information sharing among users. Lin et al. \cite{10079087} proposed a blockchain-assisted SC framework for AI generated content services to address security issues arising from malicious semantic data transmission in SC. Guo et al. \cite{10177738} introduced a semantic importance-aware communication scheme using pre-trained language models to quantify the semantic importance of data frames, thereby reducing semantic loss in communication. 


The above works mainly studied how to use GAI to improve the structures of the SC model, but do not consider the training scheme of an efficient SC model on MUs. The traditional centralized learning method requires the MUs to transmit the local data to the central server for centralized training, which may lead to a high consumption of communication energy and a high risk of information leakage \cite{rao2024privacy}. Hence, this traditional approach is not suitable for MUs to train the SC models.

Federated Learning (FL) \cite{yang2019federated} has the potential to alleviate the above issues. FL enables several clients and a central server to train the SC models collaboratively only by sharing model parameters, rather than transmitting large raw training data. 
Numerous studies have focused on communication-efficient FL. For example, Nguyen et al. \cite{9829327} presented a high-compression FL scheme that effectively reduced data load during FL processes without modifying structure or hyperparameters. Similarly, Wang et al. \cite{pmlr-v162-wang22o} proposed a communication-efficient adaptive federated optimization method that substantially lowered communication costs through error feedback and compression strategies. 
Furthermore, H{\"o}nig et al. \cite{pmlr-v162-honig22a} developed a doubly-adaptive quantization FL algorithm that dynamically adjusted the quantization level over time and among various clients, enhancing compression while maintaining model quality.
Although these studies present efficient FL algorithms to enhance model training, they ignore the issues of model adaptation for heterogeneous MUs and the high communication overheads in dynamic networks.

Based on the above review of related work, we summarize three critical challenges that apply SC to MUs as follows:
\begin{enumerate}[]
	\item {\it{Insufficient semantic extraction capabilities}:} 
	We assume that the data type transmitted between MUs and BS is the image, thus we should deploy an image SC system on them.
	Although Convolutional Neural Networks (CNNs) exhibit excellent capabilities in representing local features of images, they are difficult to capture global information effectively \cite{liu2022high}. Consequently, image semantic encoders and decoders constructed based on CNNs can not simultaneously account for both global and local semantic features of images.
	
	\item {\it{Model adaptation for heterogeneous devices in FL}:} 
	MUs are usually heterogeneous, which means they have different scales of local data and computation resources. Generally, more complex models can achieve higher accuracy when the data and computation resources are enough \cite{o2019deep}. Hence, the MUs, having more available data and computation resources, may need sophisticated models to achieve higher accuracy. However, the limited-resource MUs can only use a compact model for local training, and the well-resourced MUs have to choose the same compact model as a compromise to meet the model isomorphism requirement of FL.
	
	\item {\it{High communication overheads in dynamic networks:}} While conventional FL algorithms facilitate distributed training using local data from multiple MUs, ensuring data privacy and security, they often result in substantial network traffic and communication overhead due to frequent parameter exchanges \cite{chen2021communication}. Communication-efficient FL methods, such as those in \cite{pmlr-v162-wang22o} and \cite{pmlr-v162-honig22a}, alleviate communication energy consumption by compressing transmitted parameters. However, these compression techniques are performed on the client side, leading to additional costs for clients. Moreover, these methods do not account for the impact of fluctuating network conditions, such as variations in Signal-to-Noise Ratio (SNR) in wireless communications. 
\end{enumerate}

In this paper, we propose a novel GAI-assisted SC (GSC) model to apply in the communications between MUs and BS, improving the utilization of limited spectrum resources. Then, we present a Personalized Semantic Federated Learning (PSFL) to train the GSC models on MUs while protecting privacy and security. 
The main contributions are summarized as follows:
\begin{enumerate}[]
	\item {\it{Accurate semantic transmission through GSC}}:
	Considering the shortcomings of CNNs, we employed Vision Transformer (ViT) networks in both the semantic encoder and decoder in the GSC model. As a common GAI network for processing images, ViT can achieve more accurate semantic feature extraction of transmitted images at the transmitter and more precise image reconstruction at the receiver through the multi-head self-attention mechanism. 
	As a result, we introduce the GSC model as the communication bridge between MUs and the BS, achieving accurate semantic transmission. 
	Therefore, we solve the proposed first challenge.
	
	\item {\it{High-quality local training in PSFL}}:
	We propose a Personalized Local Distillation (PLD) strategy in the local training phase of PSFL, improving the accuracy of the GSC model. In PLD, each MU can select a suitable GSC model as a mentor according to their local resources and a unified CNN-based SC (CSC) model as a student.
	Then, the mentor model can be distilled to the student model to meet the model isomorphism requirement of the FL. As a result, PLD addresses the second challenge.
	
	\item {\it{Energy-efficent global aggregation in PSFL}}: 
	We design an Adaptive Global Pruning (AGP) algorithm in the global aggregation phase of PSFL, reducing the consumption of communication energy. Specifically, we perform pruning on the aggregated global FL model (i.e., the updated CSC model). The pruning ratio is determined by considering the real-time SNR between MUs and the BS. Therefore, the AGP solves the last challenge.
\end{enumerate}

The rest of this paper can be organized as follows. The system model is introduced in Section II. The proposed PSFL is described in Section III. The complexity analysis is introduced in Section IV. Numerical results are presented in Section V. The work summary and future expectations are described in Section VI.

\section{System Model}
\begin{figure}[htbp]
	\centering
	\includegraphics[width=8.5cm]{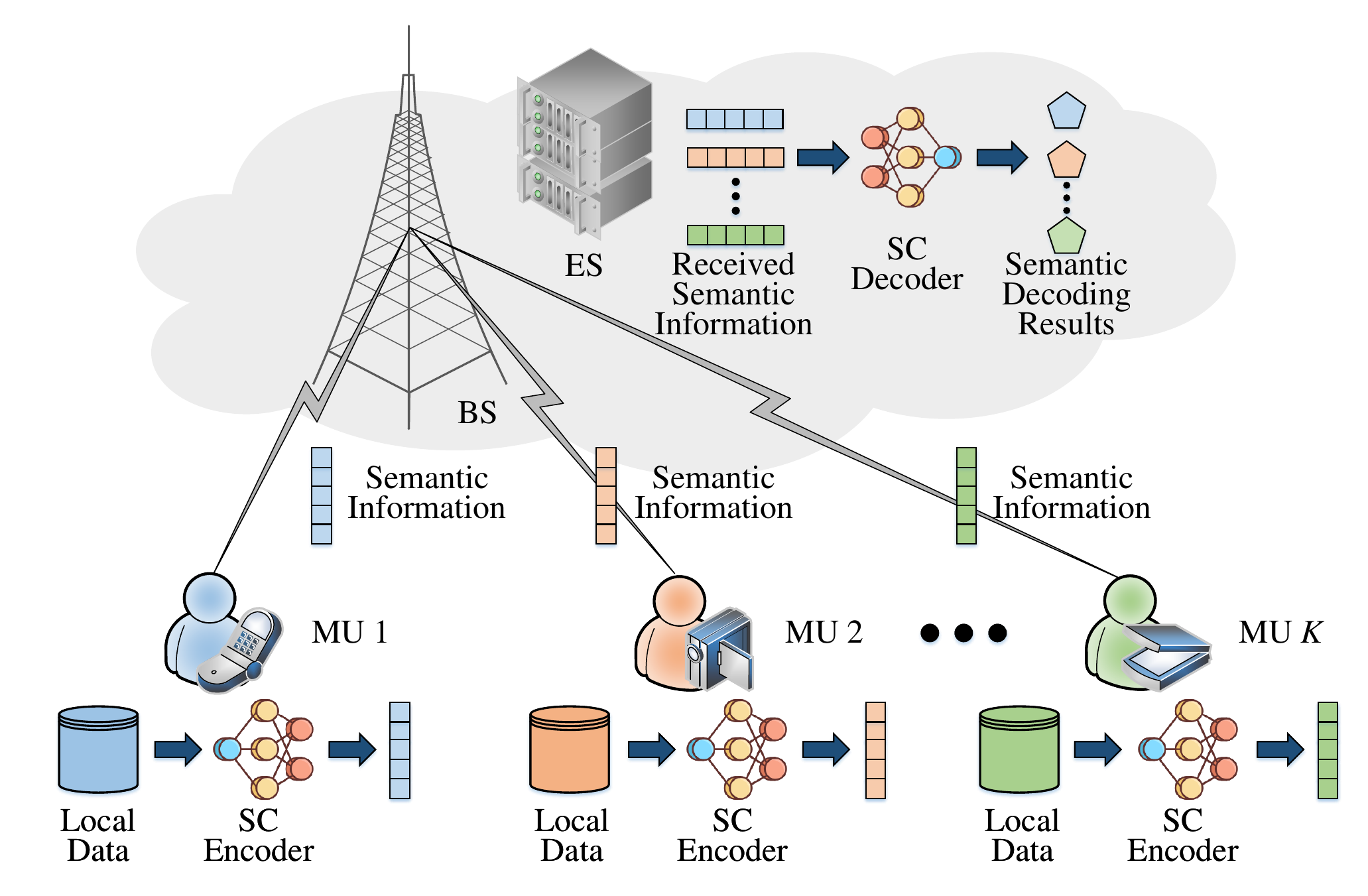}
	\caption{The illustration that MUs communicate with BS using SC.}
	\label{fig:system}
\end{figure}

Fig. \ref{fig:system} illustrates the communication between MUs and the BS through the SC system. We consider an uplink wireless network with limited spectrum resources to deploy a distributed SC system, comprising \(K\) MUs, denoted by the set \(\mathcal{K}\), and a single BS with an ES. 
In the training phase, the ES is responsible for performing global aggregation and updating the global SC model, while the MUs train their local SC models based on local data and subsequently transmit the model parameters to the BS. 
In the inference phase, the MUs only need to transmit semantic information to the BS during data transmission, rather than the large-sized raw data \cite{xie2020lite}. This semantic information is then decoded at the ES. To facilitate the extraction of semantic information, the SC encoder is deployed on each MU, while the SC decoder is deployed on the ES to decode the received semantic information.
Additionally, we consider the impairments of the physical channels between the MUs and the BS.

\subsection{GSC Model}
\begin{figure*}[htbp]
	\centering
	\includegraphics[width=15cm]{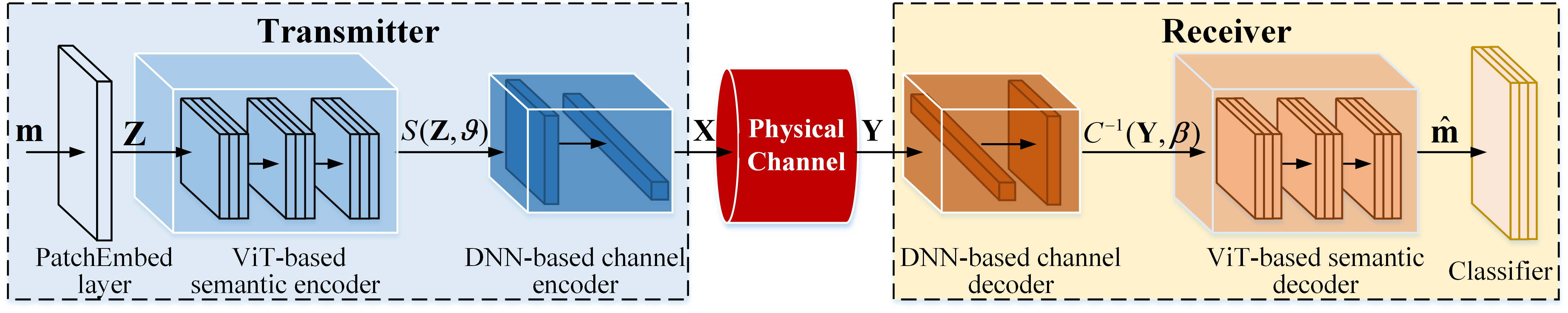}
	\caption{The illustration of image transmission utilizing the proposed GSC model.}
	\label{fig:gsc}
\end{figure*}

We mainly consider the image SC which refers to capturing the semantics of interest in input images, thereby reducing the amount of data required for image transmission and conserving bandwidth.
Compared to traditional CNNs, ViT displays superior feature analysis capabilities in various visual tasks, such as image classification, object detection, and feature extraction \cite{jiang2024large3d}. Therefore, as shown in Fig. \ref{fig:gsc}, the GSC model employs ViT as the image semantic encoder and decoder. Subsequently, based on Deep Neural Networks (DNNs), we construct the channel encoder and decoder. Finally, we apply a perception model to simulate the physical channel, ensuring it supports backpropagation. The transmission process of image SC based on the GSC model is as follows:

\subsubsection{Transmitter}
First, the input image \(\mathbf{m} \in \mathbb{R}^{H \times W \times C}\) is inputted into a PatchEmbed layer, in which $\mathbf{m}$ is converted into \(N\) patches of size \((P^2 \times C)\), where \(H\) denotes the height of the image, \(W\) denotes the width, \(C\) denotes the number of channels, and \(P^2\) represents the number of segments into which the image is divided. Hence, \(N = \frac{W \times H}{P^2}\). 

Then, the sequence \(X_p\) composed of these \(N\) patches undergoes the Patch Embedding operation \cite{dosovitskiy2020image}. Specifically, each patch in \(X_p\) bypasses a linear transformation, reducing the dimensionality of the sequence to \(D\) and resulting in a linear embedding sequence:
\begin{equation}
	\mathbf{Z}_0 = \left[X_p^1E; X_p^2E; \ldots; X_p^N E\right],
\end{equation}
where \(E\) is the linear transformation (i.e., a fully connected layer) with input dimensions \((P^2 \times C)\) and output dimensions \(N\). \(X_p^i\) represents the \(i\)-th patch in \(X_p\).

Next, the position vectors from the positional encoding, which contain positional information, are linearly combined with $\mathbf{Z}_0$ to obtain the input sequence of ViT as follows:
\begin{equation}
	\mathbf{Z} = \left[X_p^1E + P_1; X_p^2E + P_2; \ldots; X_p^N E + P_N\right],
\end{equation}
where \(P_i\) denotes the \(i\)-th position vector.
Subsequently, the semantic encoder extracts features from \(\mathbf{Z}\), resulting in the semantic feature of the original image \(\mathbf{m}\). The image semantic encoder is based on ViT, with its core being the multi-head attention layer which can learn the relationships between pixels through the multi-head attention mechanism, enabling more accurate feature extraction and realistic image reconstruction. 
The multi-head attention layer is essentially composed of multiple self-attention heads concatenated together. A self-attention head is derived from a single self-attention layer, which can be calculated by:
\begin{equation}
	head_i = \operatorname{Attention}(\mathbf{Q}, \mathbf{K}, \mathbf{V}) = \operatorname{softmax} \left( \frac{\mathbf{QK}^T}{\sqrt{d_k}} \right) \mathbf{V}, 
\end{equation}
where \(\mathbf{Q}\) is the query vector, \(\mathbf{K}\) is the matching vector corresponding to \(\mathbf{Q}\), and \(\mathbf{V}\) is the information vector. \(\mathbf{Q}\), \(\mathbf{K}\), and \(\mathbf{V}\) are all obtained through linear transformations of the input \(\mathbf{Z}\), i.e., \(\mathbf{Q} = \mathbf{Z}\mathbf{W}_\mathbf{Q}\), \(\mathbf{K} = \mathbf{Z}\mathbf{W}_\mathbf{K}\), and \(\mathbf{V} = \mathbf{Z}\mathbf{W}_\mathbf{V}\), where \(\mathbf{W}_\mathbf{Q}\), \(\mathbf{W}_\mathbf{K}\), and \(\mathbf{W}_\mathbf{V}\) are the respective weight matrices. \(d_k\) is the scaling factor.
Assuming the number of heads is \( h \), the multi-head self-attention can be calculated by:
\begin{equation}
	\begin{split}
		&\operatorname{MultiheadAttention}(\mathbf{Q}, \mathbf{K}, \mathbf{V}) =\\
		&\text{concat}(head_1, \cdots, head_h)\mathbf{W}_\text{mha},
	\end{split}
\end{equation}
where \(\mathbf{W}_\text{mha}\) represents the weights of the multi-head attention layer.

Finally, to ensure data is transmitted on the physical channel, the semantic feature should be encoded and modulated by the channel encoder to reduce channel impairments and improve robustness. The transmitted signal can be represented as:
\begin{equation}
	\mathbf{X} = C\left(S\left(\mathbf{Z},\bm{\vartheta}\right),\bm{\alpha}\right),
\end{equation}
where $S\left(\cdot\right)$ represents the semantic encoder with model parameters $\bm{\vartheta}$ and $C\left(\cdot\right)$ is the channel encoder with model parameters $\bm{\alpha}$. 

\subsubsection{Physical channel}
When transmitted over the physical channel, $\mathbf{X}$ suffers transmission impairments that include attenuation and noise.  
The transmission process of the physical channel can be expressed as:
\begin{equation}\label{eq:Shi2}
	\mathbf{Y} = \mathbf{H \cdot X+N},
\end{equation}
where $\mathbf{Y}$ represents the received signal; $\mathbf{H}$ represents the channel gain between the transmitter and the receiver; $\mathbf{N}$ is Additive White Gaussian Noise (AWGN). For end-to-end training of encoder and decoder, the physical channel must allow back-propagation \cite{xie2021deep}, hence we use a perception model to simulate the physical channel.

\subsubsection{Receiver}
The channel decoder demodulates the received signal to extract the semantic features, which are then decoded by the semantic decoder. The recovered image, \(\hat{\mathbf{m}}\), is obtained as follows:
\begin{equation}\label{eq:Shi3}
	\hat{\mathbf{m}}=S^{-1}\left(C^{-1}(\mathbf{Y}, \bm{\beta}), \bm{\delta}\right),
\end{equation}
where $C^{-1}\left(\cdot\right)$ represents the channel decoder with model parameters $\bm{\beta}$; $S^{-1}\left(\cdot\right)$ is the semantic decoder with model parameters $\bm{\delta}$.

In this paper, we consider the classification tasks-oriented SC. Hence, we use a classifier to perform image classification based on $\mathbf{\hat{m}}$ and the cross-entropy (CE) as the loss function of the GSC model:
\begin{equation}\label{eq:Shi4}
	\mathcal{L}_\text{CE}(\mathbf{y}, \mathbf{\hat{y}})=-\sum_{i=1}^{M} y_{i} \log \left(\hat{y}_{i}\right),
\end{equation}
where $\mathbf{y}=[y_1,y_2,..., y_M]$ represents the labels, when the input data $m$ belongs to the $i$-th class, $y_i=1$, otherwise 0; $\mathbf{\hat{y}}=[\hat{y}_1,\hat{y}_2,..., \hat{y}_M]$ represents the predicted probabilities based on $\hat{\mathbf{m}}$, $\hat{y}_i$ represents the probability predicted as the $i$-th class; $M$ is the total number of categories.

\subsection{FL Model}
We define the local dataset of the $k$-th MU as  $\mathcal{D}_k=\left\{ (\mathbf{m}_{k,1},y_{k,1}), (\mathbf{m}_{k,2},y_{k,2}), ..., (\mathbf{m}_{k,N_k},y_{k,N_k})\right\}$, where $N_k$ is the number of collected samples in $\mathcal{D}_k$, $\mathbf{m}_{k,i}$ is the $i$-th sample and $y_{k,i}$ is the corresponding label. Note that $\mathcal{D}_k$ may be non-Independent Identical Distributed (non-IID) data which depends on the realistic environment and usage pattern of the $k$-th MU. 

For the $k$-th MU, the local loss function in the $t$-th communication round of FL can be calculated as:
\begin{align}\label{eq:Shi5}
	{F_{k}}\left( \mathbf{w}_{k,t} \right) = \frac{1}{N_k}{\sum\limits_{i=1}^{N_k}{f\left( \mathbf{w}_{k,t}, \mathbf{m}_{k,i},y_{k,i}\right)}},
\end{align}
where $f(\mathbf{w}_{k,t},\mathbf{m}_{k,i},y_{k,i})$ is the training loss function for the $i$-th sample $\left(\mathbf{m}_{k,i},y_{k,i} \right)$ in $\mathcal{D}_{k}$; $\mathbf{w}_{k,t}$ is the weights of the local FL model of the $k$-th MU in the $t$-th communication round. $\mathbf{w}_{k,t}$ includes all the parameters of the GSC model, namely $\mathbf{w}_{k,t}=(\bm{\alpha}_{k,t},\bm{\vartheta}_{k,t},\bm{\beta}_{k,t}, \bm{\delta}_{k,t})$, where $\bm{\alpha}_{k,t},\bm{\vartheta}_{k,t},\bm{\beta}_{k,t},\bm{\delta}_{k,t}$ represent the parameters of the GSC model deployed on the $k$-th MU from the channel encoder to the semantic decoder in the $t$-th communication round.

To ensure data privacy and security, the global FL model is updated in ES by aggregating the local FL models from MUs, the update of the global FL model can be given by \cite{yang2019federated}:
\begin{align}\label{eq:Shi6}
	\mathbf{w}_{g,t}=\frac{\sum\limits_{k=1}^{K} N_{k} \mathbf{w}_{k,t}}{\sum\limits_{k=1}^{K} N_{k}},
\end{align}
where $\mathbf{w}_{g,t}$ is the weights of the global FL model in the $t$-th communication round. Note that $\mathbf{w}_{g,t}$ shares the same architecture to $\mathbf{w}_{k,t}$. In addition, since the transmitted parameters in FL may also leak sensitive data, as a solution, the differential privacy and homophobic encryption algorithms can be used to encrypt the transmitted parameters and improve the data security and privacy \cite{qu2022blockchain}.

FL aims to get the optimal global model $\mathbf{w}_{g,t}$ that can minimize the local FL loss of all devices, so as to achieve global optimization. Hence, the global loss function of FL can be given by:
\begin{equation}\label{eq:Shi7}
	F_{g}\left( \mathbf{w}_{g,t} \right) = \frac{1}{K}\sum\limits_{k = 1}^{K}{F_{k}\left(\mathbf{w}_{g,t}\right)},
\end{equation}
where $F_{k}\left(\mathbf{w}_{g,t}\right)$ represents that the local FL loss based on $\mathbf{w}_{g,t}$. In this paper, we take minimizing $F_{g}\left( \mathbf{w}_{g,t} \right)$ as the goal of training GSC model. 

\subsection{Communication Model}
We consider that Orthogonal Frequency Division Multiple Access (OFDMA) is adopted for the links between MUs and the BS. When the $k$-th MU uploads its local model weights $\mathbf{w}_{k,t}$ to BS, the uplink rate can be given by:
\begin{equation}\label{eq:vku}
	v_{k,t}= B_{k,t} \log _{2}\left(1+\psi_{k,t}\right),
\end{equation}
where $B_{k,t}$ and $\psi_{k,t}$ represent the uploading bandwidth and SNR of the $k$-th MU in the $t$-th communication round, respectively. The transmission delay between the $k$-th MU and the BS over uplink in the $t$-th communication round is calculated as:
\begin{equation}\label{eq:tku}
	\tau_{k,t}=\frac{Z\left(\mathbf{w}_{k,t}\right)}{v_{k,t}},
\end{equation}
where $Z(\mathbf{w}_{k,t})$ represents the number of bits that each MU $k$ requires transmitting to the BS. The energy consumption of the communication process can be given by:
\begin{equation}\label{eq:eku}
	E_{k,t}^\text{com}=P_{k,t}\tau_{k,t},
\end{equation}
where $P_{k,t}$ is the transmitting power of the $k$-th MU in the $t$-th communication round. The communication energy of FL is another critical optimization goal.

\section{Personalized Semantic Federated Learning for GSC Model}
To address the challenges of training GSC models deployed on MUs, we propose the PSFL, in which we apply the PLD strategy and AGP algorithm to optimize FL in the phases of local training and global aggregation, respectively. 
\begin{figure*}[htpb]
	\centering
	\includegraphics[width=16cm]{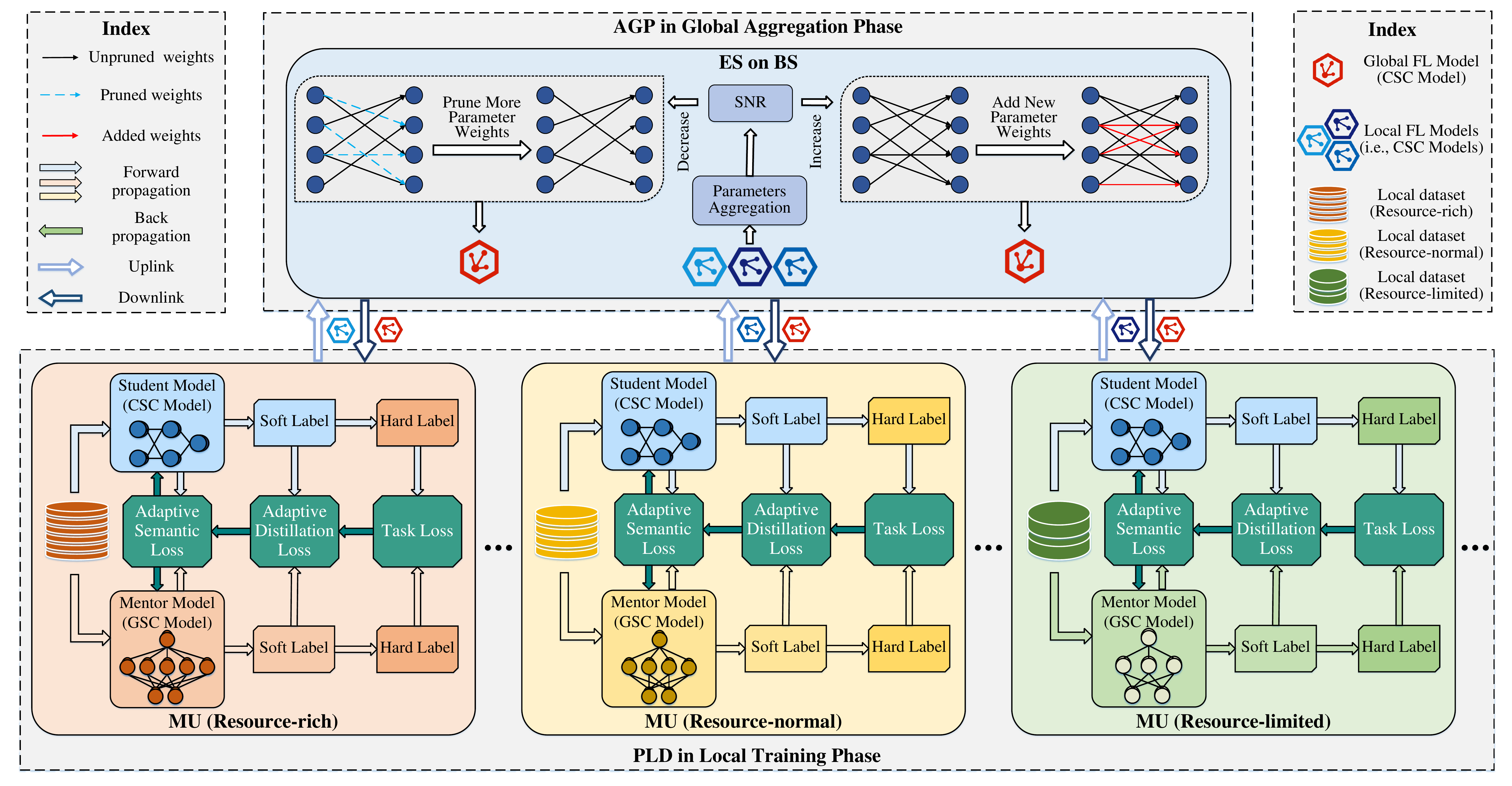}
	\caption{The illustration of the proposed PSFL.}
	\label{fig:FL}
\end{figure*}
\subsection{PLD for Local Training}
To address the issue of model adaptation for heterogeneous MUs and ensure effective information exchange among different GSC models, we propose the PLD strategy during the local training phase. Specifically, as shown in Fig. \ref{fig:FL}, each MU not only deploys a suitable GSC model based on their local resources but also deploys a unified and small-scale CSC model.
The GSC model is used for SC services after training. The CSC model, serving as a vehicle for transferring knowledge of the GSC model, is uploaded to the BS for parameter aggregation. It is then returned to the local MU, transmitting the newly aggregated knowledge back to the GSC model. This process indirectly facilitates the information exchange of heterogeneous GSC models across different MUs.
To achieve effective knowledge exchange between the GSC and CSC models in this process, Knowledge Distillation (KD) is utilized. 

KD is a transfer learning method involving a sophisticated mentor model and a compact student model, aiming to transfer knowledge from the mentor to the student model \cite{shen2020federated}. In PLD, the GSC model acts as the mentor while the CSC model is the student. In PLD, the mutual learning process between the mentor and student models based on KD is as follows:

\subsubsection{Distill knowledge from hard labels}
The mentor and student models compute the loss between the output of models and hard labels \cite{jafari2021annealing}. Generally, the hard labels are determined by the specific task. Since we consider the classification task, the hard labels are the categories of the input data. We denote the input data on the $k$-th MU as $\mathbf{m}_k$, and the corresponding hard labels as $\mathbf{y}_k$. The communication loss functions for classification tasks of mentor and student models are expressed as follows:
\begin{equation}\label{eq:ltm}
\mathcal{L}_\text{task}^{\prime}=\mathcal{L}_\text{CE}(\mathbf{y}_k, \mathbf{\hat{y}}_k^{\prime}) + \operatorname{MSE}(\mathbf{m}_k,\mathbf{\hat{m}^\prime}_k),
\end{equation}
\begin{equation}\label{eq:lts}
\mathcal{L}_\text{task}=\mathcal{L}_\text{CE}(\mathbf{y}_k, \mathbf{\hat{y}}_k) + \operatorname{MSE}(\mathbf{m}_k,\mathbf{\hat{m}}_k),
\end{equation}
where $\mathbf{\hat{y}}_k^{\prime}$ and $\mathbf{\hat{y}}_k$ represent the probabilities of $\mathbf{m}_k$ as predicted by the mentor and student models, respectively. Similarly, $\mathbf{\hat{m}^\prime}_k$ and $\mathbf{\hat{m}}_k$ denote the reconstructed images generated by the mentor and student models after wireless communications, respectively. The function $\rm{MSE}(\cdot)$ represents the mean-square error, which is employed to ensure consistency between the original and reconstructed images at the pixel level. In summary, the task losses provide direct task-specific supervision for the mentor and student models.

\subsubsection{Distill knowledge from soft labels}
The mentor and student models transfer knowledge from the soft labels (e.g., $\mathbf{\hat{y}}_k^{\prime}$ and $\mathbf{\hat{y}}_k$) reciprocally\cite{jafari2021annealing}. Since incorrect predictions from the mentor/student model may mislead the other one in the KD, we propose an adaptive method to weigh the distillation loss according to the quality of predicted soft labels (i.e., Eqs. (\ref{eq:ltm}) and (\ref{eq:lts})). The adaptive distillation losses of mentor and student models are formulated as follows:
\begin{equation}\label{eq:ldm}
\mathcal{L}_\text{dis}^{\prime}=\frac{{\rm{KL}}(\mathbf{\hat{y}}_k^{\prime},\mathbf{\hat{y}}_k)}{\mathcal{L}_\text{task}},
\end{equation}
\begin{equation}\label{eq:lds}
\mathcal{L}_\text{dis}=\frac{{\rm{KL}}(\mathbf{\hat{y}}_k,\mathbf{\hat{y}}_k^{\prime})}{\mathcal{L}_\text{task}^{\prime}},
\end{equation}
where ${\rm{KL}}(\cdot)$ means the Kullback–Leibler divergence. In this way, the distillation intensity is weak if the predictions of the mentor and student models are not reliable (e.g., their task losses are large). The distillation loss becomes dominant if the mentor and student models are well-tuned, which means small task losses. Thus, the adaptive distillation losses have the potential to mitigate the risk of overﬁtting.

\subsubsection{Distill knowledge from the semantic information}
To improve the performance of SC, the mentor and student models learn to minimize the difference between the output of the semantic encoder and the output of the channel decoder. Similarly, to avoid the misguiding of the mentor/student models in the interactive process, we also weight the semantic loss according to task losses. Therefore, the adaptive semantic losses for the mentor and student models are formulated as follows:
\begin{equation}\label{eq:lhms}
\mathcal{L}_\text{sem}^{\prime}=\mathcal{L}_\text{sem}=\frac{{\rm{MSE}}(\mathbf{S}_k^{\prime},\mathbf{S}_k)+{\rm{MSE}}(\mathbf{C}_k^{\prime},\mathbf{C}_k)}{\mathcal{L}_\text{task}^{\prime}+\mathcal{L}_\text{task}}, 
\end{equation}
where $\mathbf{S}_k^{\prime}$ and $\mathbf{S}_k$ represent the semantic encodings of the mentor and student models, respectively; $\mathbf{C}_k^{\prime}$ and $\mathbf{C}_k$ represent the channel decodings of the mentor and student models, respectively. 

\subsubsection{Update mentor and student models}
According to the above-proposed loss functions, the total loss functions for updating the mentor and student models are formulated as follows:
\begin{equation}\label{eq:lttm}
\mathcal{L}^{\prime}_\text{total}=\mathcal{L}^{\prime}_\text{task}+\mathcal{L}^{\prime}_\text{dis}+\mathcal{L}^{\prime}_\text{sem},
\end{equation}
\begin{equation}\label{eq:ltts}
\mathcal{L}_\text{total}=\mathcal{L}_\text{task}+\mathcal{L}_\text{dis}+\mathcal{L}_\text{sem}.
\end{equation}

The mentor and student models update their weights by minimizing the total losses with the Stochastic Gradient Descent (SGD) optimizer \cite{zhang2018improved}.
We assume the training is performed on the $k$-th MU with the dataset $\mathcal{D}_k$, the weights of the mentor and student models are denoted as $\mathbf{w}_k^{\prime}$ and $\mathbf{w}_k$, respectively. $G$ is used to denote the total epochs of the local training. The entire workflow of PLD is summarized in \textbf{Algorithm \ref{alg:PLD}}.  

\begin{algorithm}
\caption{PLD}
\label{alg:PLD}
\begin{algorithmic}[1]
	\REQUIRE $G$, $\mathcal{D}_k$.
	\ENSURE $\mathbf{w}_k^{\prime}$, $\mathbf{w}_k$.
	\FOR{each epoch in $G$}
	\FOR{$i~ = ~1,~2,...,~N_k$}
	\STATE{Sample $m_{k,i}$ from $\mathcal{D}_{k}$.}
	\STATE{Compute task losses $\mathcal{L}_\text{task}^{\prime}$ and $\mathcal{L}_\text{task}$ according to Eqs. (\ref{eq:ltm}) and (\ref{eq:lts}).}
	\STATE{Compute adaptive distillation losses $\mathcal{L}_\text{dis}^{\prime}$ and $\mathcal{L}_\text{dis}$ according to Eqs. (\ref{eq:ldm}) and (\ref{eq:lds}).}
	\STATE{Compute adaptive semantic losses $\mathcal{L}_\text{sem}^{\prime}$ and $\mathcal{L}_\text{sem}$ according to Eq. (\ref{eq:lhms}).}
	\STATE{Compute total losses $\mathcal{L}_\text{total}^{\prime}$ and $\mathcal{L}_\text{total}$ according to Eqs. (\ref{eq:lttm}) and (\ref{eq:ltts}).}
	\STATE{Update $\mathbf{w}_k$ by minimizing $\mathcal{L}_\text{total}$.}
	\STATE{Update $\mathbf{w}_k^{\prime}$ by minimizing $\mathcal{L}_\text{total}^{\prime}$.}
	\ENDFOR
	\ENDFOR
\end{algorithmic}
\end{algorithm}

The advantages of PLD can be summarized as follows: (1) PLD allocates a sophisticated GSC model for each MU in the local training phase. Simultaneously, a unified and compact CSC model is used as a vehicle for transferring knowledge of the GSC model in FL, ensuring the effective information exchange between heterogeneous GSC models; (2) PLD weights the distillation and semantic losses according to the prediction results, which avoids the problem that the mentor/student model may mislead the other one in the mutual knowledge transfer.

\subsection{AGP for Global Aggregation}
In wireless environments, the traditional FL may bring unaffordable communication energy consumption for MUs due to the frequent parameter exchanges.
From above Eqs. (\ref{eq:vku})-(\ref{eq:eku}), we can see that the smaller $Z(\mathbf{w}_{k,t})$ and larger $\psi_{k,t}$ mean lower communication energy. Namely, when $\psi_{k,t}$ increases, we can transmit more parameters, otherwise we should transmit fewer parameters, thereby ensuring low consumption of communication energy in dynamic SNR. Hence, we design the AGP algorithm, in which the global FL model $\mathbf{w}_{g,t}$ (i.e., the CSC model) is adaptively pruned in the ES after global aggregation. The proportion of pruning is determined by assessing the real-time SNR between MUs and the BS. 
As shown in Fig. \ref{fig:FL}, we assume that PSFL starts from a certain round $t$, and the workflow of the  PSFL assisted by AGP is described as follows:

\subsubsection{Model pruning and weights broadcasting}
We perform pruning on the global FL model $\mathbf{w}_{g,t}$ in the ES, in which a proportion of the smallest positive weights and the largest negative weights of $\mathbf{w}_{g,t}$ will be pruned.
To avoid impairing the knowledge carried by the global FL model due to excessive pruning of parameters, we propose that the pruning proportion should adaptively adjust based on the real-time SNR between MUs and the BS in each communication round. 
Consequently, the pruning proportion in the $t$-th communication round can be represented as:
\begin{equation}\label{eq:prr}
\zeta_t = \frac{\psi_{\max}-\frac{1}{K}\sum\limits_{k \in \mathcal{K}}\psi_{k,t}}{\psi_{\max}-\psi_{\min}}
,
\end{equation}
where $\psi_{\max}$ and $\psi_{\min}$ represent the maximum and minimum SNR between the MUs and BS, respectively. 
Thereafter, BS broadcasts the pruned global FL model $\hat{\mathbf{w}}_{g,t}$ to all MUs for the $(t+1)$-th round training.

\subsubsection{Local training and weights uploading}
Each MU performs local training with the PLD strategy and obtains the latest mentor and student models. Since the FL model is pruned, it can be denoted as $\hat{\mathbf{w}}_{k,t+1}$. Afterward, the local FL model $\hat{\mathbf{w}}_{k,t+1}$ is uploaded to BS by the wireless channel.

\subsubsection{Global aggregation and model updating}
The local FL models from all MUs are aggregated according to Eq. (\ref{eq:Shi6}) in ES. Afterward, according to the variation of the SNR between the MUs and BS, we calculate the new pruning proportion $\zeta_{t+1}$. If $\zeta_{t+1} < \zeta_{t}$, more weights of the global FL model should be pruned, otherwise we add weights randomly until $\zeta_{t+1} = \zeta_{t}$. Finally, the pruned global FL model $\hat{\mathbf{w}}_{g,t+1}$ is downloaded to each MU to update their local FL models:
\begin{equation}\label{eq:flup}
\hat{\mathbf{w}}_{k,t+1}=\hat{\mathbf{w}}_{g,t+1}.
\end{equation}

Assuming the total number of communication rounds is $T$, the proposed PSFL assisted by AGP is summarized in \textbf{Algorithm \ref{alg:AGP}}.

\begin{algorithm}
\caption{PSFL assisted by AGP}
\label{alg:AGP}
\begin{algorithmic}[1]
	\REQUIRE $T$, $\psi_{k,t}$.
	\ENSURE $\mathbf{\hat{w}}_{g,T}$.
	\FOR{each communication round $t \in T$}
	\STATE{\textbf{BS do}}
	\STATE{Aggregate local FL models from MUs according to Eq. (\ref{eq:Shi6}).}
	\STATE{Calculate the pruning proportion $\zeta_t$ by Eq.  (\ref{eq:prr}).}
	\STATE{Prune the global FL model $\mathbf{w}_{g,t}$ and obtain $\mathbf{\hat{w}}_{g,t}$.}
	\STATE{Broadcast the pruned global FL model $\hat{\mathbf{w}}_{g,t}$ to each device to update the local FL model according to Eq. (\ref{eq:flup}).}
	\STATE{\textbf{Each MU do}}
	\STATE{Train the mentor and local FL models by the PLD strategy in \textbf{Algorithm \ref{alg:PLD}}.}
	\STATE{Upload the latest local FL model $\hat{\mathbf{w}}_{k,t+1}$ to the BS.}
	\ENDFOR
\end{algorithmic}
\end{algorithm}

The novelty of the presented AGP method can be summarized as follows:
(1) \emph{Adaptive pruning in dynamic environments}:
AGP algorithm can increase the pruning rate to ensure stable transmission and low communication cost for bad communication situations, and it can also recover pruned parameters to improve the learning ability for good communication situations.
(2) \emph{Trade-off between accuracy and energy consumption}:
AGP algorithm can achieve an elaborate trade-off between model accuracy and energy consumption in the dynamic network environment. 

\section{Complexity Analysis}
The average data size of each MU is denoted as $D=\frac{1}{K}\sum_{k \in \mathcal{K}}N_k$. We assume the complexity of communication and computation is linearly proportional to the model size \cite{wu2022communication}. Then, the complexity analysis of PSFL in terms of communication and computation is performed.

\textit{\textbf{Communication complexity analysis}}: In the traditional FL, without PLD and AGP, the mentor model performs both local training and global aggregation. Hence, the communication complexity is $O(T|\mathbf{w}_{k,t}^{\prime}|)$, where $|\cdot|$ represents the operator of calculating the size of parameter weights. In PSFL, the complexity of communication is $O(T|\hat{\mathbf{w}}_{k,t}|\zeta_t)$, which is much smaller than traditional FL for $|\hat{\mathbf{w}}_{k,t}|<|\mathbf{w}_{k,t}|\ll|\mathbf{w}_{k,t}^{\prime}|$ and $\zeta_t<1$.

\textit{\textbf{Computation complexity analysis}}: The computation complexity that directly learning the large mentor model in FL is $O(TD|\mathbf{w}_{k,t}^{\prime}|)$. In PSFL, the computation complexity consists of two parts, i.e., local mentor and FL models training, which are $O(TD|\mathbf{w}_{k,t}^{\prime}|)$ and $O(TD|\hat{\mathbf{w}}_{k,t}|)$, respectively. Hence, the computation complexity of PSFL is $O(TD|\mathbf{w}_{k,t}^{\prime}|)+O(TD|\hat{\mathbf{w}}_{k,t}|)$. Since the model size of the pruned FL model is much smaller than the mentor model, the extra computation cost of learning the FL model is
much smaller than learning the large mentor model, namely, the computation complexity of PSFL is $O(TD|\mathbf{w}_{k,t}^{\prime}|)$.

In practice, compared with the standard FedAvg algorithm, the extra computation cost of training the FL model is much smaller than training the large mentor model with the PLD strategy. AGP can also stably reduce communication energy consumption due to the adaptive prune. Thus, the proposed PSFL is much more communication-efficient than the standard FedAvg algorithm and meanwhile does not introduce much computation cost.
\section{Numerical Results}

\subsection{Simulation Settings}
Firstly, we assess the proposed PSFL scheme using the MNIST \cite{lecun1998gradient}, Fashion-MNIST \cite{xiao2017fashion}, CIFAR-10 \cite{9372789}, and CIFAR-100 \cite{huang2024fedsr} datasets.
In the MNIST and Fashion-MNIST datasets, the training set contains 60,000 samples and the testing set contains 10,000 samples, distributed across 10 categories.
The CIFAR-10 dataset comprises 50,000 RGB images as training samples and 10,000 RGB images as test samples, distributed across 10 categories.
The CIFAR-100 dataset comprises 50,000 RGB images as training samples and 10,000 RGB images as test samples, distributed across 100 categories.
Furthermore, we apply a Dirichlet distribution \cite{li2021model} to generate the non-IID data partition among MUs, where the concentration parameter of the Dirichlet distribution is denoted as $r$ and set to 0.9 by default. Thus, each device can have relatively balanced data samples of some classes.

Secondly, we assume the participation of 9 MUs as clients in the training process. The maximum transmit power is configured as $P_{k,\max}=0.1$ W. 
The maximum and minimum SNR between the MUs and BS are set as $\psi_{\max}=25$ dB and $\psi_{\min} = 0$ dB, respectively.
We utilize the AWGN channel as the simulated physical channel for the GSC model to better demonstrate its ability to resist channel interference. Additionally, the SNRs between the BS and clients fluctuate across different communication rounds, ranging from 0 dB to 25 dB.

Thirdly, Masked Autoencoders (MAE) \cite{he2022masked} is a well-known large generative model, which utilizes ViT as the encoder-decoder to learn accurate semantic representation of images and perform high-quality construction of images.
Therefore, we adopt the MAE to construct the semantic encoder and decoder in the GSC model, respectively.
As shown in TABLE \ref{tab:exp}, we consider three kinds of GSC models and denote them as GSC-M, GSC-L, and GSC-H according to their parameter amount. 
In the PSFL scheme, the three kinds of GSC models are used as the mentor models.
In the CSC model used as the student, we adopt the ResNet-18 \cite{rezende2017malicious} as the semantic encoder and construct the semantic decoder using three layers of deconvolutional network.
As a result, each client is equipped with a unified student model and a personalized mentor model based on the size of their local datasets.
\renewcommand{\arraystretch}{1.5} 
\begin{table*}[htbp]
	\centering
	\caption{Data distribution and local model settings for each client}
	\label{tab:exp}
	\begin{tabular}{|c|ccccccccc|}
		\hline
		Index of client   & \multicolumn{1}{c|}{0}    & \multicolumn{1}{c|}{1}    & \multicolumn{1}{c|}{2}    & \multicolumn{1}{c|}{3}    & \multicolumn{1}{c|}{4}    & \multicolumn{1}{c|}{5}    & \multicolumn{1}{c|}{6}    & \multicolumn{1}{c|}{7}    & 8    \\ \hline
		Local data volume & \multicolumn{1}{c|}{1854} & \multicolumn{1}{c|}{3703} & \multicolumn{1}{c|}{4429} & \multicolumn{1}{c|}{4783} & \multicolumn{1}{c|}{5467} & \multicolumn{1}{c|}{5634} & \multicolumn{1}{c|}{5891} & \multicolumn{1}{c|}{8958} & 9281 \\ \hline
		Mentor model      & \multicolumn{4}{c|}{GSC-M}                                                                                    & \multicolumn{3}{c|}{GSC-L}                                                        & \multicolumn{2}{c|}{GSC-H}       \\ \hline
		Parameters  & \multicolumn{4}{c|}{112,442,112}                                                                              & \multicolumn{3}{c|}{330,287,872}                                                  & \multicolumn{2}{c|}{657,924,684} \\ \hline
		Student model     & \multicolumn{9}{c|}{CSC}                                                                                                                                                                                                       \\ \hline
		Parameters   & \multicolumn{9}{c|}{54,721,065}                                                                                                                                                                                                           \\ \hline
	\end{tabular}
	
\end{table*}

Finally, our simulations are performed with the PyTorch framework on a server, which has an Intel Xeon CPU (2.4 GHz, 128 GB RAM) and an NVIDIA A800 GPU (80 GB SGRAM).

\subsection{Evaluation of the proposed PSFL}
In this subsection, we aim to evaluate the performance of the proposed PSFL scheme in terms of both loss and accuracy. 
Note that accuracy refers to the probability of correctly classifying an image reconstructed by the GSC model using a pre-trained classifier network. We employ ResNet-101 \cite{jusman2023comparison} as the pre-trained classifier.
Additionally, the loss and accuracy results shown in the figures represent the mean loss and accuracy of the GSC model across all clients.

\begin{figure*}[htbp]
	\centering
	\includegraphics[width=16cm]{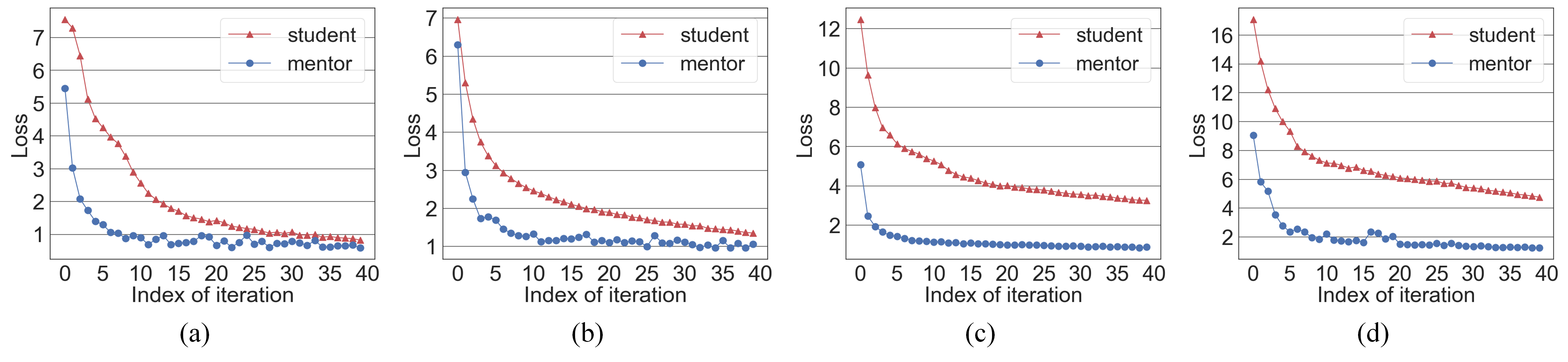}
	\caption{Loss versus iteration under student and mentor models on datasets (a) MNIST, (b) Fashion-MNIST, (c) CIFAR-10, and (d) CIFAR-100.}
	\label{fig:exp1}
\end{figure*}

\begin{figure*}[htbp]
	\centering
	\includegraphics[width=16cm]{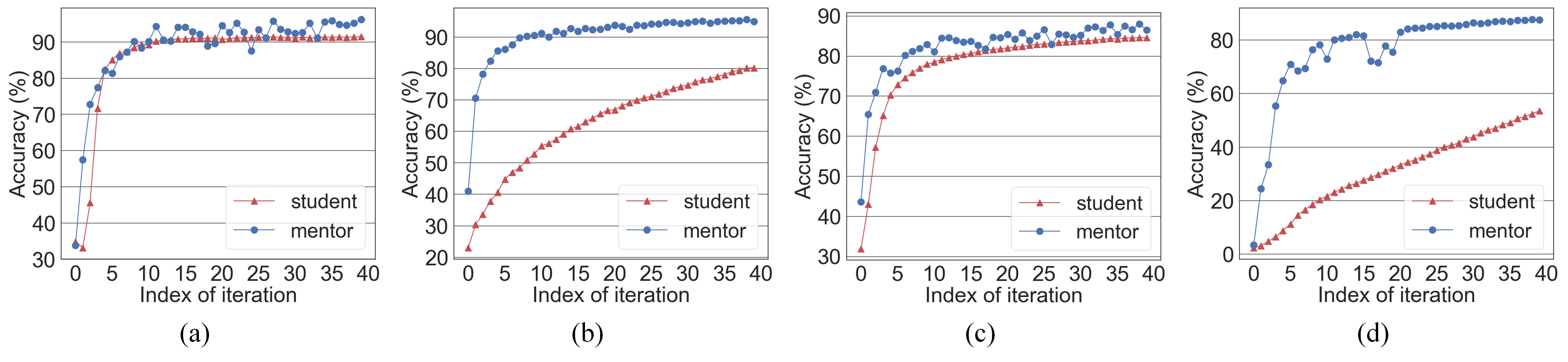}
	\caption{Accuracy versus iteration under student and mentor models on datasets (a) MNIST, (b) Fashion-MNIST, (c) CIFAR-10, and (d) CIFAR-100.}
	\label{fig:exp2}
\end{figure*}

Firstly, Fig. \ref{fig:exp1} and Fig. \ref{fig:exp2} illustrate the training results of the mentor and student models on four datasets.
We can observe that the loss and accuracy of the mentor are still higher than those of the student model, which illustrates the strong guidance ability of the mentor model. The mentor models always have the ability to direct the learning of the student models.
Specifically, as shown in Fig. \ref{fig:exp1}(a)-(b) and Fig. \ref{fig:exp2}(a)-(b), for the performances of the mentor and student models on the MNIST and Fashion-MNIST datasets, due to the simplicity of these two datasets, the student model performs as well as the mentor model. This suggests that they are learning from each other in the training process.
However, as shown in Fig. \ref{fig:exp1}(c)-(d) and Fig. \ref{fig:exp2}(c)-(d), the CIFAR-10 and CIFAR-100 datasets are more complex; hence, the more sophisticated mentor model outperforms the student model. This means the mentor model can provide direction for the student during the training process.
This evaluation demonstrates that in the PSFL scheme, both the mentor and student models can continue to learn and improve, ensuring the effectiveness of information exchange.

\begin{figure}[htbp]
	\centering
	\includegraphics[width=8.5cm]{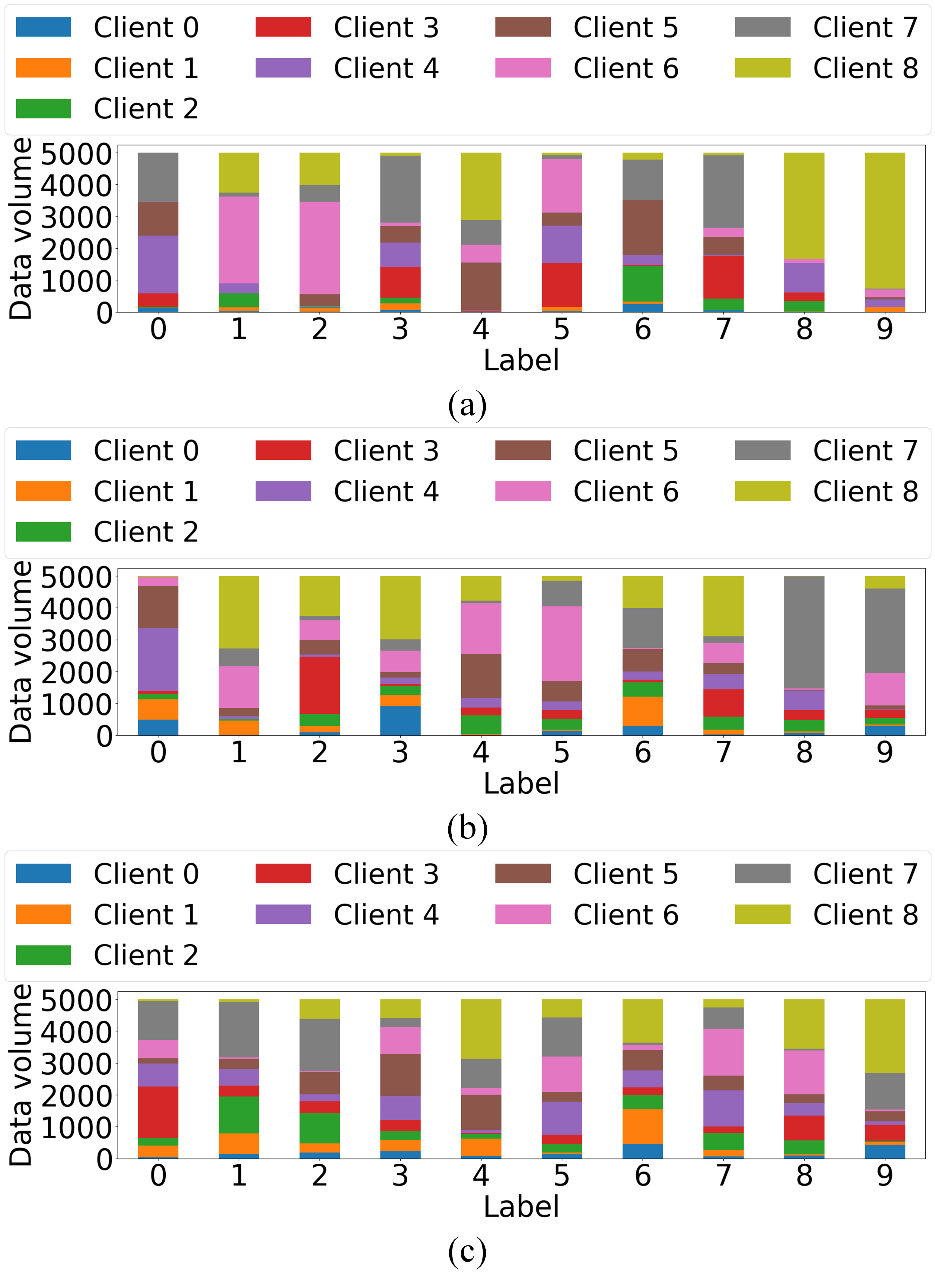}
	\caption{Data distribution of each client on CIFAR-10 dataset when (a) $r=0.3$, (b) $r=0.6$, and (c) $r=0.9$, respectively.}
	\label{fig:datadist}
\end{figure}

\begin{figure*}[htbp]
	\centering
	\includegraphics[width=16cm]{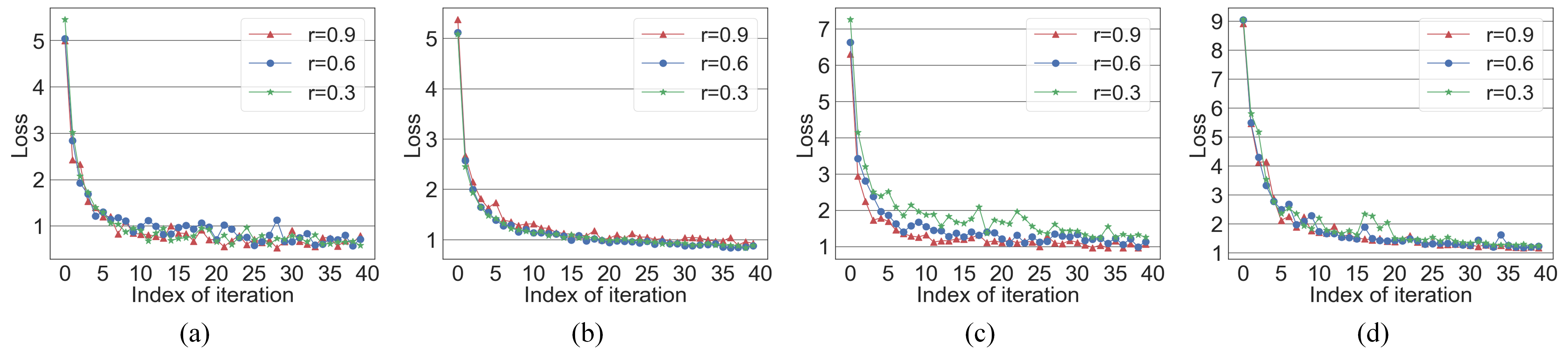}
	\caption{Loss versus iteration under different concentration parameters $r$ of the Dirichlet distribution on datasets (a) MNIST, (b) Fashion-MNIST, (c) CIFAR-10, and (d) CIFAR-100.}
	\label{fig:exp3}
\end{figure*}

\begin{figure*}[htbp]
	\centering
	\includegraphics[width=16cm]{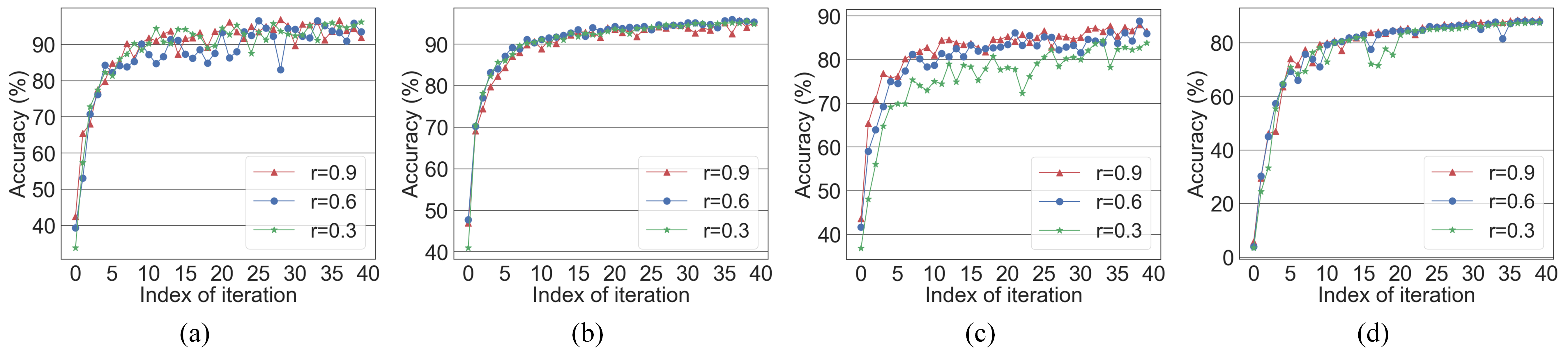}
	\caption{Accuracy versus iteration under different concentration parameters $r$ of the Dirichlet distribution on datasets (a) MNIST, (b) Fashion-MNIST, (c) CIFAR-10, and (d) CIFAR-100.}
	\label{fig:exp4}
\end{figure*}

Secondly, we evaluate the model performance under different $r$.
To more intuitively illustrate the impact of $r$, Fig .\ref{fig:datadist} shows the distribution of different categories of data in the CIFAR-10 dataset on each client.
Fig. \ref{fig:exp3} and Fig. \ref{fig:exp4} illustrate the training results under different Dirichlet distributions on four datasets. 
The results suggest that the performance of the model becomes worse with the decrease in $r$, as a smaller $r$ results in greater differences between different client data. Specifically, as shown in Fig. \ref{fig:exp3}(a)-(b) and Fig. \ref{fig:exp4}(a)-(b), for the simpler MNIST and Fashion-MNIST datasets, the differences are smaller under varying $r$. For the more complex CIFAR-10 and CIFAR-100 datasets, as shown in Fig. \ref{fig:exp3}(c)-(d) and Fig. \ref{fig:exp4}(c)-(d), the increase in $r$ has a more significant impact on the FL model. This means that the proposed PSFL can not perform well with non-IID data when the dataset is complex, which could lead to further improvements in the FL scheme in the future.

\begin{figure*}[htbp]
	\centering
	\includegraphics[width=16cm]{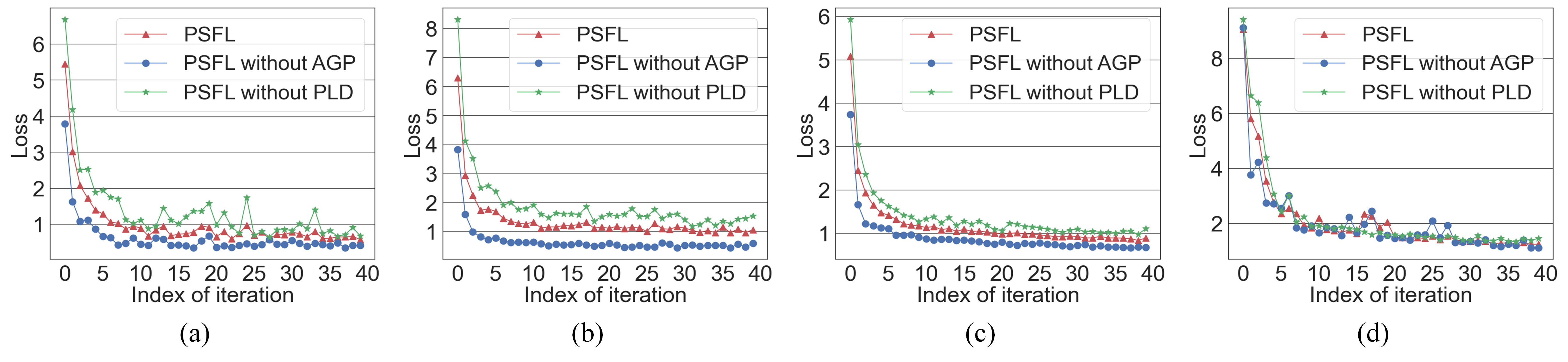}
	\caption{Loss versus iteration under different methods on datasets (a) MNIST, (b) Fashion-MNIST, (c) CIFAR-10, and (d) CIFAR-100.}
	\label{fig:exp5}
\end{figure*}

\begin{figure*}[htbp]
	\centering
	\includegraphics[width=16cm]{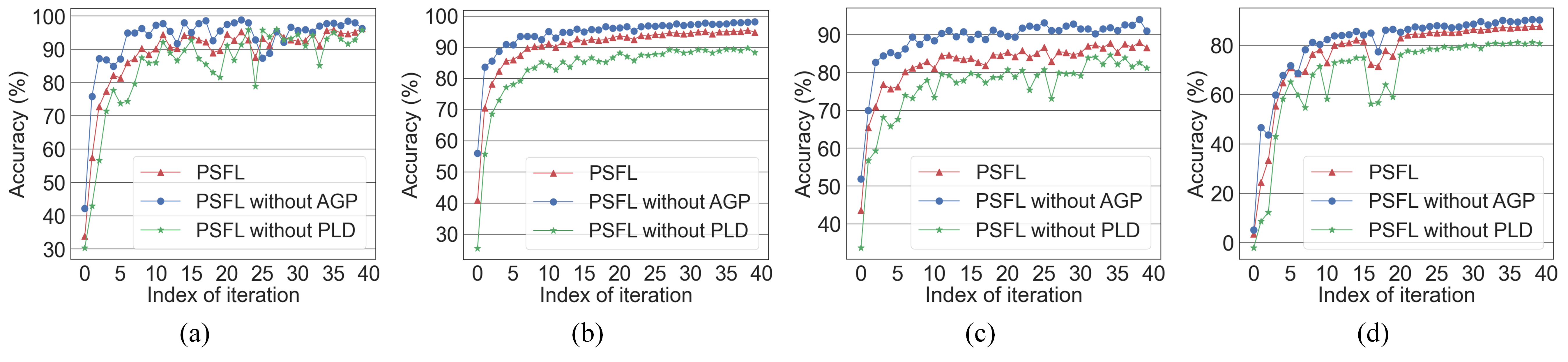}
	\caption{Accuracy versus iteration under different methods on datasets (a) MNIST, (b) Fashion-MNIST, (c) CIFAR-10, and (d) CIFAR-100.}
	\label{fig:exp6}
\end{figure*}

Lastly, to evaluate the functions of PLD and AGP in the PSFL scheme, we performed ablation experiments, and the results are shown in Fig. \ref{fig:exp5} and Fig. \ref{fig:exp6}. Note in the PSFL without PLD, each MU selects the GSC-M model as the mentor.
From Fig. \ref{fig:exp5}(a)-(b) and Fig. \ref{fig:exp6}(a)-(b), we can see that the PSFL without AGP, PSFL, and PSFL without PLD schemes have similar performance. 
This may be due to the simplicity of the MNIST and Fashion-MNIST datasets. In Fig. \ref{fig:exp5}(c)-(d) and Fig. \ref{fig:exp6}(c)-(d), the PSFL without AGP achieves the lowest loss and the highest accuracy, while the PSFL without PLD has the worst result. We can speculate that AGP reduces the parameters of the model but impacts the performance, whereas PLD effectively improves model accuracy. Additionally, Fig. \ref{fig:exp6}(c)-(d) shows that the difference between PSFL without AGP and PSFL is smaller than that between PSFL and PSFL without PLD. This illustrates that AGP has less impact on model accuracy than PLD.

\subsection{Evaluation of different contenders}
In this subsection, we compare the proposed PSFL to the other FL schemes in terms of global loss, global accuracy, and local accuracy.
The following methods are introduced in this experiment as contenders:
\begin{itemize}
	\item FedAvg: A common FL approach, which is equivalent to the PSFL without PLD and AGP algorithms \cite{yang2019federated}.
	\item STC: A compressed FL framework that is designed to meet the requirements of the FL environment \cite{sattler2019robust}.
	\item FTTQ: A parameter quantization-based communication-efﬁcient FL approach \cite{xu2020ternary}.
	\item FedPAQ: A communication-efficient FL method with periodic averaging and quantization \cite{reisizadeh2020fedpaq}.
	\item PSFL: The proposed FL approach in this paper.
\end{itemize}

\begin{figure*}[htbp]
	\centering
	\includegraphics[width=16cm]{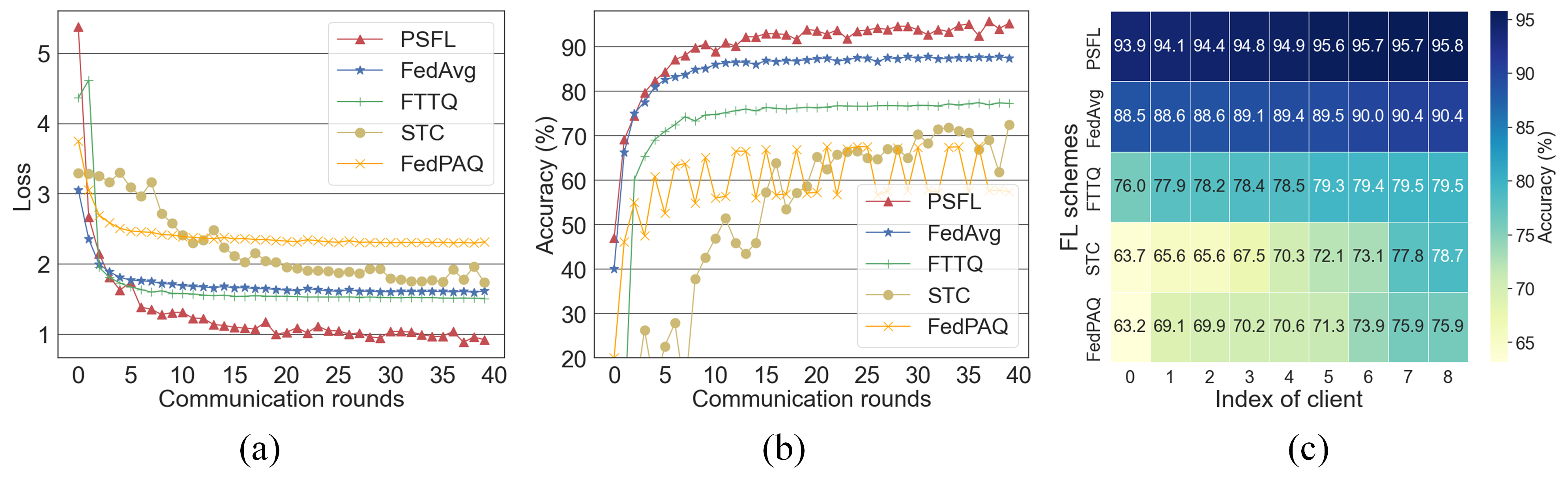}
	\caption{Comparison results of different schemes on Fashion-MNIST in terms of (a) global loss, (b) global accuracy, and (c) local accuracy of each client.}
	\label{fig:exp7}
\end{figure*}

\begin{figure*}[htbp]
	\centering
	\includegraphics[width=16cm]{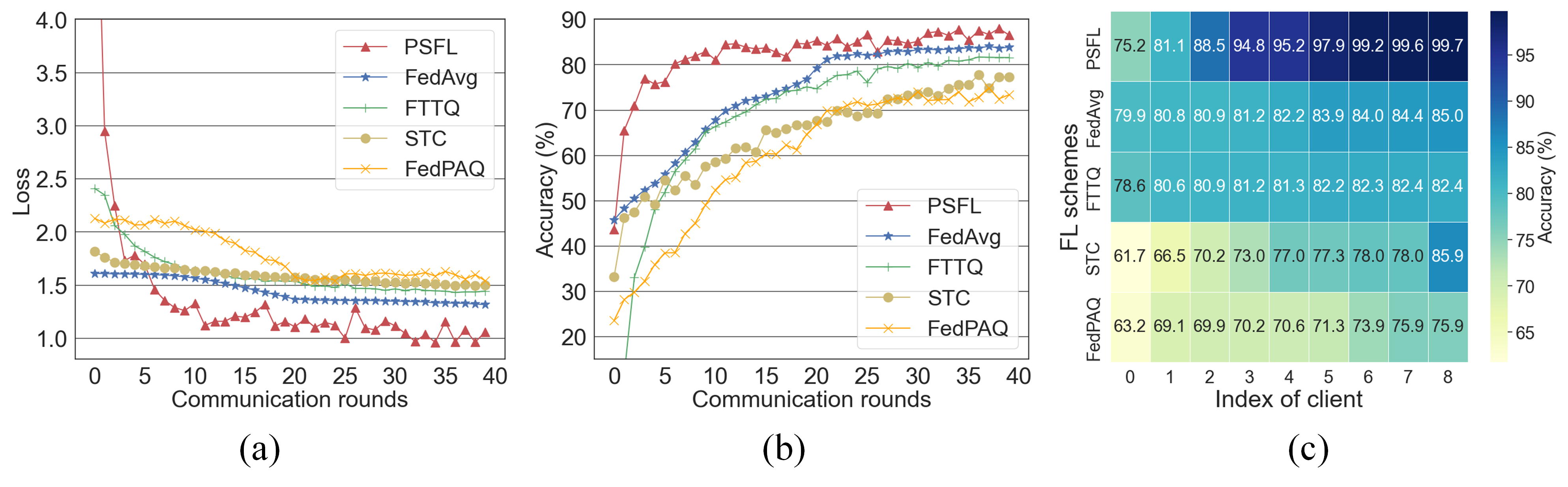}
	\caption{Comparison results of different schemes on CIFAR-10 in terms of (a) global loss, (b) global accuracy, and (c) local accuracy of each client.}
	\label{fig:exp8}
\end{figure*}

Except for PSFL, the other schemes adopt the GSC-M as the FL model. Additionally, for simplicity, we only evaluate these methods on the Fashion-MNIST and CIFAR-10 datasets, and we set $r=0.9$. Fig. \ref{fig:exp7} and Fig. \ref{fig:exp8} show evaluation results.

The loss results in Fig. \ref{fig:exp7}(a) and Fig. \ref{fig:exp8}(a) suggest that the proposed PSFL could converge to the best point on the Fashion-MNIST and CIFAR-10 datasets, while FedPAQ performs the worst. Additionally, the FedAvg and FTTQ schemes perform better, while STC performs worse.
From Fig. \ref{fig:exp7}(b) and Fig. \ref{fig:exp8}(b), we can see that the accuracy of the global FL model obtained by our method is the best and is significantly better than that of the other contenders. The FedAvg method performs better than the FTTQ, STC, and FedPAQ methods, with FedPAQ performing the worst.
In Fig. \ref{fig:exp7}(c) and Fig. \ref{fig:exp8}(c), we can see that the proposed PSFL enables all local FL models to achieve the best final accuracy. The performance of FedAvg and FTTQ is only slightly worse than ours, while STC performs only slightly better than FedPAQ, with FedPAQ results being the worst on both datasets.

We speculate that the excellent accuracy performance of the PSFL is mainly attributed to the PLD strategy. PLD allows clients to freely select the most compatible GSC models, thus fully utilizing available resources and achieving high accuracy. Hence, with the proposed PSFL, all clients could achieve the best performance despite differences in their local data resources. Moreover, without transmitting all parameter weights, the AGP algorithm also ensures effective model information exchange among clients, thereby maintaining the accuracy of the GSC models.

\subsection{Performance Evaluation of Communication Energy Consumption}
\begin{figure}[htbp]
	\centering
	\includegraphics[width=8.5cm]{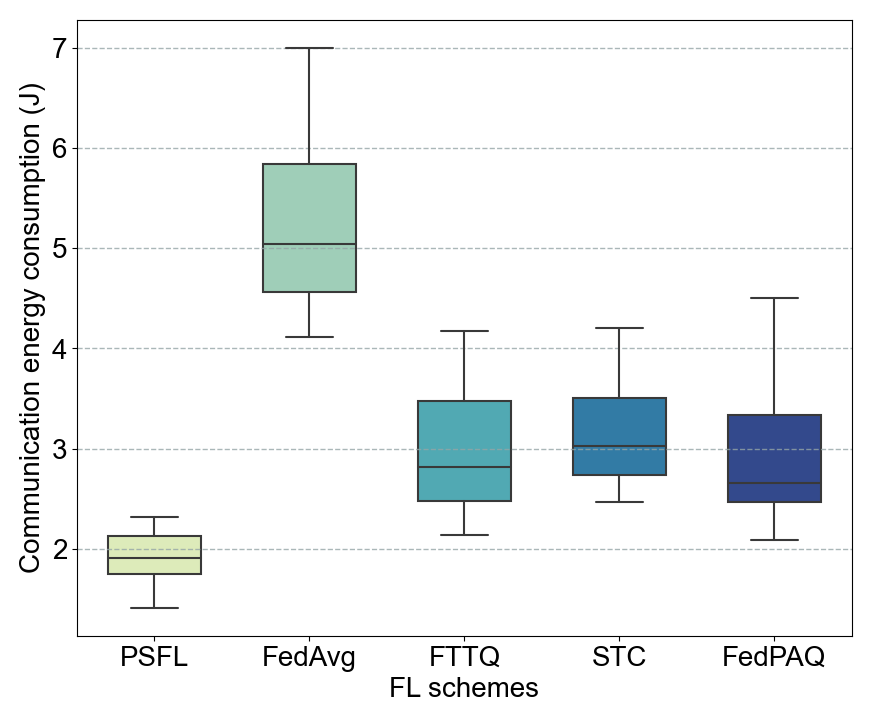}
	\caption{Communication energy consumption of each communication round using different methods.}
	\label{fig:exp9}
\end{figure}

In this subsection, we evaluate the performance of the proposed PSFL and other schemes in terms of communication energy consumption. Fig. \ref{fig:exp9} shows the communication energy consumption of each communication round using different FL schemes.

From Fig. \ref{fig:exp9}, we can see that the boxplot of the proposed PSFL is at the bottom, which means the communication energy consumption of PSFL is the lowest. Meanwhile, the boxplot of PSFL is the flattest, namely, the energy consumption change of each round is the smallest under dynamic SNR. Similarly, we can see the highest energy consumption and the most dramatic variation of energy consumption in FedAvg. 

Hence, we demonstrate the proposed PSFL can ensure low communication energy consumption in dynamic SNR.
The low and stable communication energy consumption of the PSFL can be attributed to AGP. The AGP algorithm prunes the FL model while considering the dynamic SNR, thus reducing the communication energy consumption efficiently in wireless communications. Furthermore, the pruning operator is processed only on the server, which could not bring extra cost to clients.

\section{Conclusion}
In this paper, we propose a novel GSC model that leverages the strengths of GAI to enhance the performance of SC between MUs and the BS. Additionally, we introduce the PSFL framework to enable MUs and the BS to collaboratively train GSC models while accommodating the training requirements of heterogeneous MUs. In PSFL, we first introduce the PLD strategy during the local training phase, in which each MU selects a suitable GSC model as a mentor and a unified CSC model as a student. The two models engage in mutual learning based on KD. After local training, the unified CSC model is utilized as the local FL model and uploaded to the BS for parameter aggregation, thereby obtaining the global FL model.
Secondly, we apply the AGP algorithm in the global aggregation phase, which prunes the aggregated global FL model according to the real-time SNR. The AGP algorithm reduces the transmitted model parameters and achieves the trade-off between the communication energy and model accuracy. Finally, numerical results demonstrate the feasibility and efficiency of the proposed PSFL. 

In the future, we will work on improving the performance of the proposed PSFL on non-IID data by introducing the latest personalized FL algorithms. 
Additionally, since the parameters of the CSC model may relate to user privacy, improving the security of model parameters during parameter aggregation is also a potential issue.

\bibliographystyle{ieeetran}
\bibliography{bare_jrnl_bobo}
\section*{Biographies}
\vspace{-10mm}
\begin{IEEEbiography}[{\includegraphics[width=1in,height=1.25in,keepaspectratio]{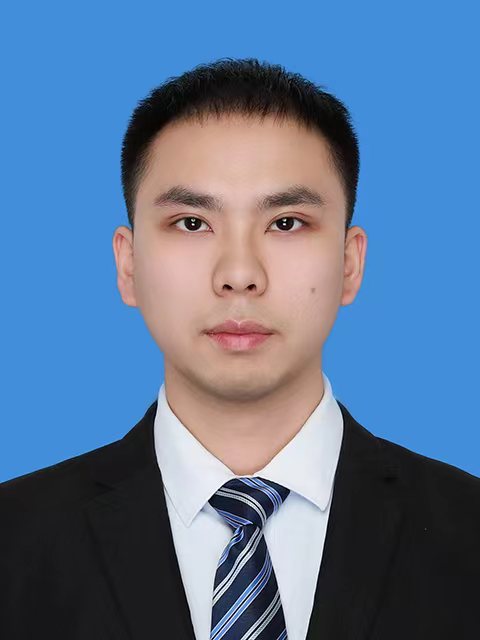}}]{Yubo Peng} received his B.S. and M.S. degrees from Hunan Normal University, Changsha, China, in 2019 and 2024. He is pursuing a doctor’s degree from the School of Intelligent Software and Engineering at Nanjing University. His main research interests include semantic communication and large models.
\end{IEEEbiography}

\begin{IEEEbiography}[{\includegraphics[width=1in,height=1.25in,keepaspectratio]{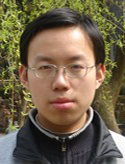}}]{Feibo Jiang} received his B.S. and M.S. degrees in School of Physics and Electronics from Hunan Normal University, China, in 2004 and 2007, respectively. He received his Ph.D. degree in School of Geosciences and Info-physics from the Central South University, China, in 2014. He is currently an associate professor at the Hunan Provincial Key Laboratory of Intelligent Computing and Language Information Processing, Hunan Normal University, China. His research interests include artificial intelligence, fuzzy computation, Internet of Things, and mobile edge computing.
\end{IEEEbiography}

\vspace{-10mm}
\begin{IEEEbiography}[{\includegraphics[width=1in,height=1.25in,keepaspectratio]{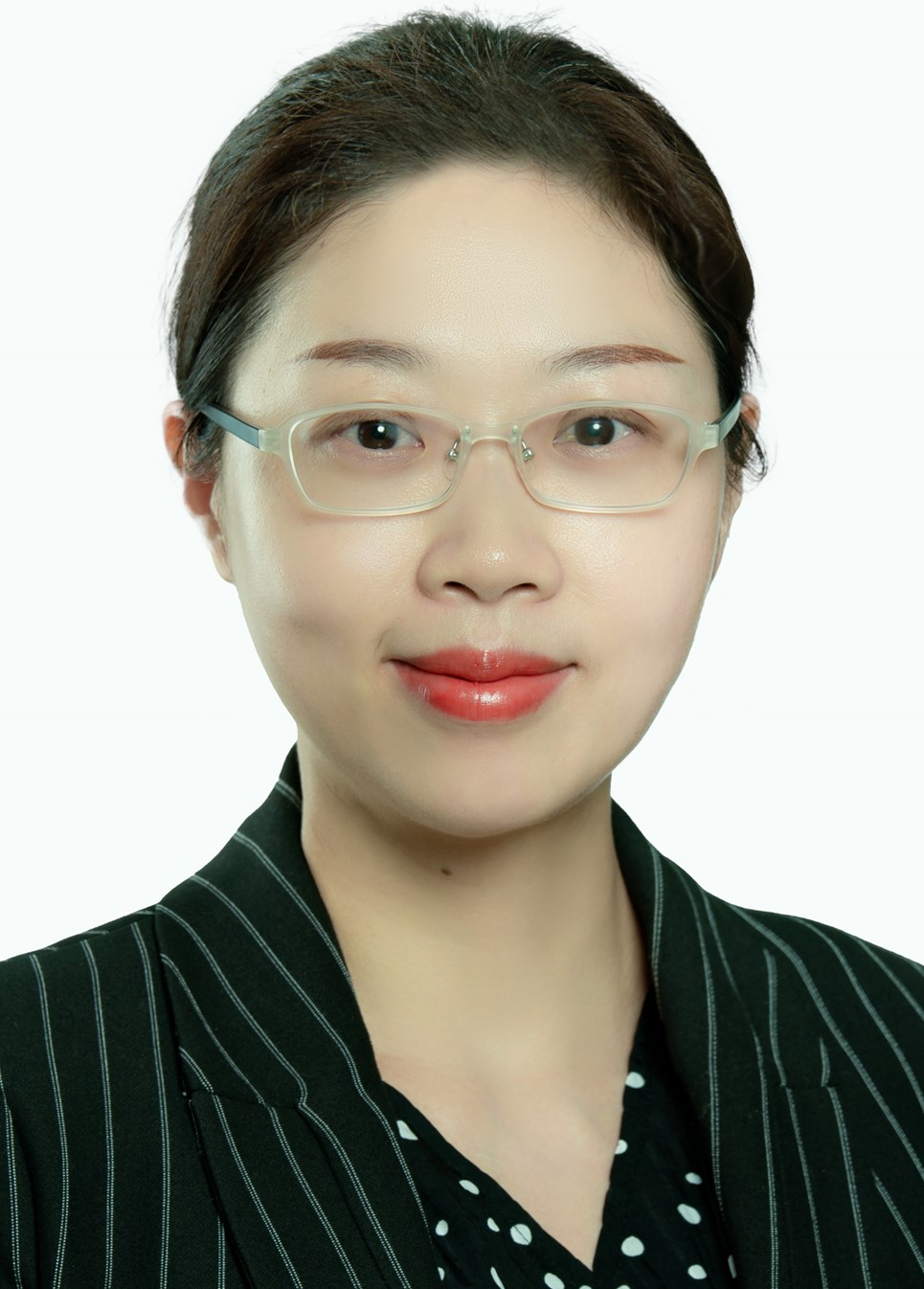}}]{Li Dong} received the B.S. and M.S. degrees in School of Physics and Electronics from Hunan Normal University, China, in 2004 and 2007, respectively. She received her Ph.D. degree in School of Geosciences and Info-physics from the Central South University, China, in 2018. Her research interests include machine learning, Internet of Things, and mobile edge computing.
\end{IEEEbiography}

\vspace{-10mm}
\begin{IEEEbiography}[{\includegraphics[width=1in,height=1.25in,keepaspectratio]{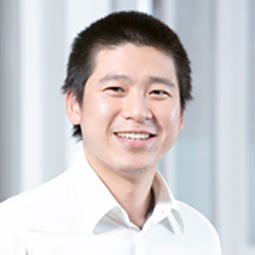}}]{Kezhi Wang}received the Ph.D. degree in Engineering from the University of Warwick, U.K. He was with the University of Essex and Northumbria University, U.K. Currently, he is a Senior Lecturer with the Department of Computer Science, Brunel University London, U.K. His research interests include wireless communications, mobile edge computing, and machine learning.
\end{IEEEbiography}

\vspace{-10mm}
\begin{IEEEbiography} [{\includegraphics[width=1in,height=1.25in,clip,keepaspectratio]{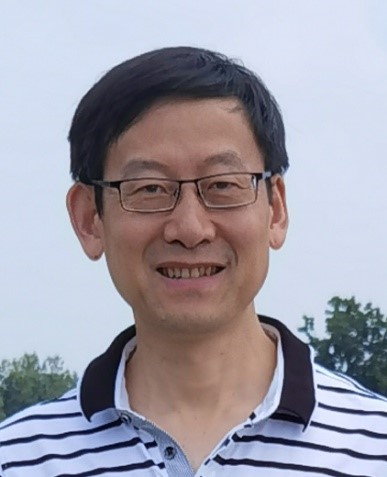}}]{Kun Yang} received his PhD from the Department of Electronic \& Electrical Engineering of University College London (UCL), UK. He is currently a Chair Professor in the School of Intelligent Software and Engineering, Nanjing University, China. He is also an affiliated professor of University of Essex, UK. His main research interests include wireless networks and communications, communication-computing cooperation, and new AI (artificial intelligence) for wireless. He has published 400+ papers and filed 30 patents. He serves on the editorial boards of a number of IEEE journals (e.g., IEEE WCM, TVT, TNB). He is a Deputy Editor-in-Chief of IET Smart Cities Journal. He has been a Judge of GSMA GLOMO Award at World Mobile Congress – Barcelona since 2019. He was a Distinguished Lecturer of IEEE ComSoc (2020-2021). He is a Member of Academia Europaea (MAE), a Fellow of IEEE, a Fellow of IET and a Distinguished Member of ACM.
\end{IEEEbiography} 
\newpage
\end{document}


%
\title{Supplementary Material}

\markboth{Submitted for Review}%
{Shell \MakeLowercase{\textit{et al.}}: Bare Demo of IEEEtran.cls for IEEE Journals}
%



\maketitle 


%
\IEEEpeerreviewmaketitle

\section{Evaluation of the proposed GSC model}
To emphasize the advantages of the GAI architecture in an intuitive way, we display some restructured images generated by the GSC and CSC models. Additionally, we employ evaluation metrics, including Peak Signal-to-Noise Ratio (PSNR) and Structural Similarity Index Measure (SSIM), to quantify the quality of the reconstructed images. PSNR measures the quality of a reconstructed image, typically expressed in decibels, with higher values indicating better image quality. The definition of PSNR is as follows:
\begin{equation}
	\mathrm{PSNR}(\mathbf{m},\mathbf{\hat{m}}) = 10 \cdot \log_{10} \left( \frac{\mathrm{MAX}_I^2}{\mathrm{MSE}(\mathbf{m},\mathbf{\hat{m}})} \right),
\end{equation}
where $\mathrm{MAX}_I$ denotes the maximum possible pixel value of the image, which is typically 255 for an 8-bit image. 
Similarly, SSIM is a metric that gauges the perceived similarity between two images, factoring in three key components - luminance, contrast, and structure. The definition of SSIM is outlined as follows:
\begin{equation}
	\mathrm{SSIM}(\mathbf{m},\mathbf{\hat{m}}) = \frac{(2\varphi_{\mathbf{m}}\varphi_{\mathbf{\hat{m}}} + c_1)(2\phi_{\mathbf{m}\mathbf{\hat{m}}} + c_2)}{(\varphi_{\mathbf{m}}^2 + \varphi_{\mathbf{\hat{m}}}^2 + c_1)(\phi_{\mathbf{m}}^2 + \phi_{\mathbf{\hat{m}}}^2 + c_2)},
\end{equation}
where $\varphi_{\mathbf{m}}$ and $\varphi_{\mathbf{\hat{m}}}$ are their means; $\phi_{\mathbf{m}}^2$ and $\phi_{\mathbf{\hat{m}^2}}$ are their variances; $\phi_{\mathbf{m}\mathbf{\hat{m}}}$ is their covariance; $c_1$ and $c_2$ are two constants used to avoid division by zero.

\begin{figure}[htbp]
	\centering
	\includegraphics[width=8.5cm]{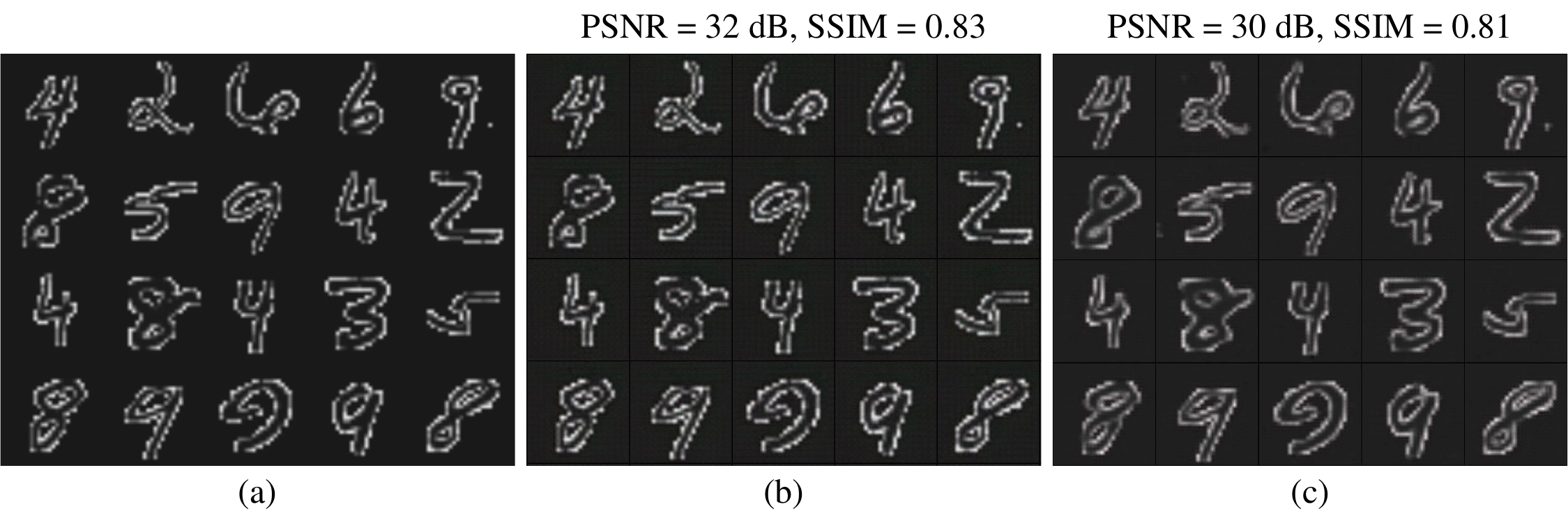}
	\caption{Image transmission results on the MNIST dataset. (a) Original images. (b) Reconstructed images based on the GSC model. (c) Reconstructed images based on the CSC model.}
	\label{fig:GSC_exp1}
\end{figure}

\begin{figure}[htbp]
	\centering
	\includegraphics[width=8.5cm]{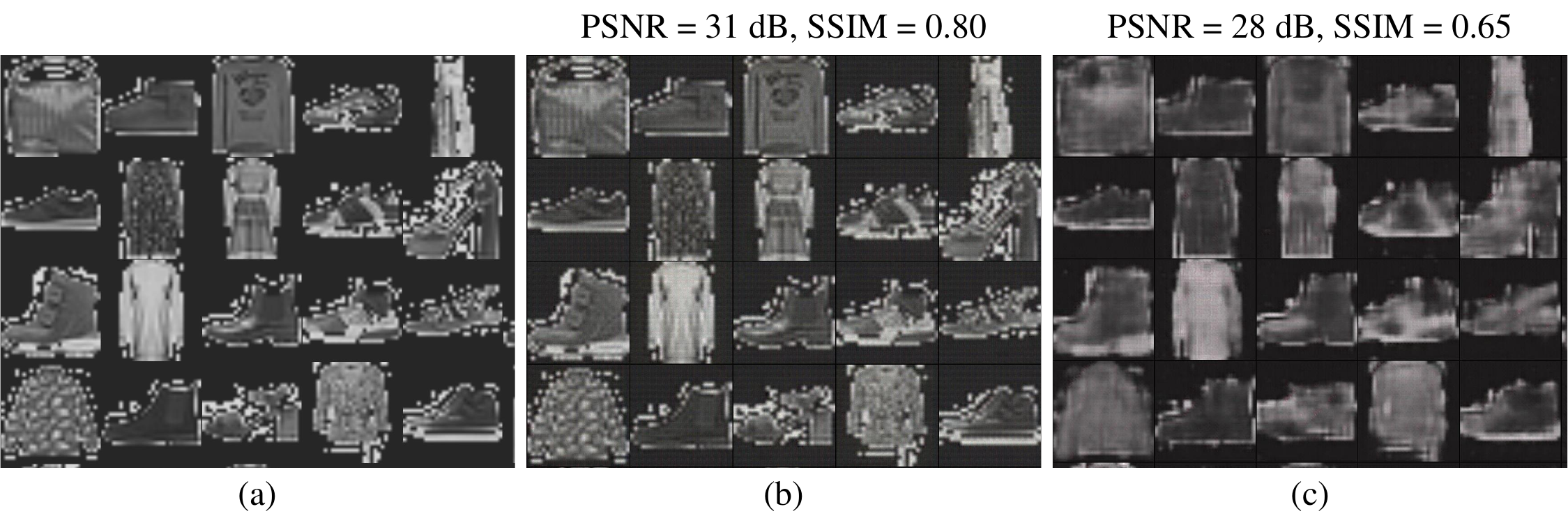}
	\caption{Image transmission results on the Fashion-MNIST dataset. (a) Original images. (b) Reconstructed images based on the GSC model. (c) Reconstructed images based on the CSC model.}
	\label{fig:GSC_exp2}
\end{figure}

\begin{figure}[htbp]
	\centering
	\includegraphics[width=8.5cm]{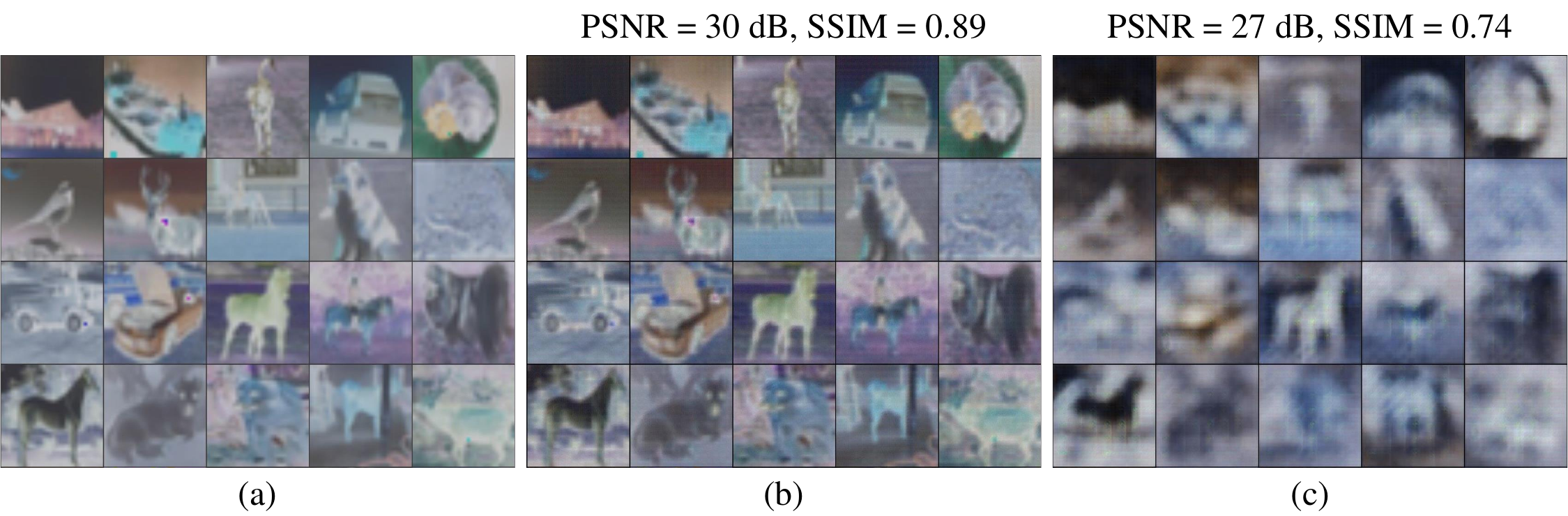}
	\caption{Image transmission results on the CIFAR-10 dataset. (a) Original images. (b) Reconstructed images based on the GSC model. (c) Reconstructed images based on the CSC model.}
	\label{fig:GSC_exp3}
\end{figure}

\begin{figure}[htbp]
	\centering
	\includegraphics[width=8.5cm]{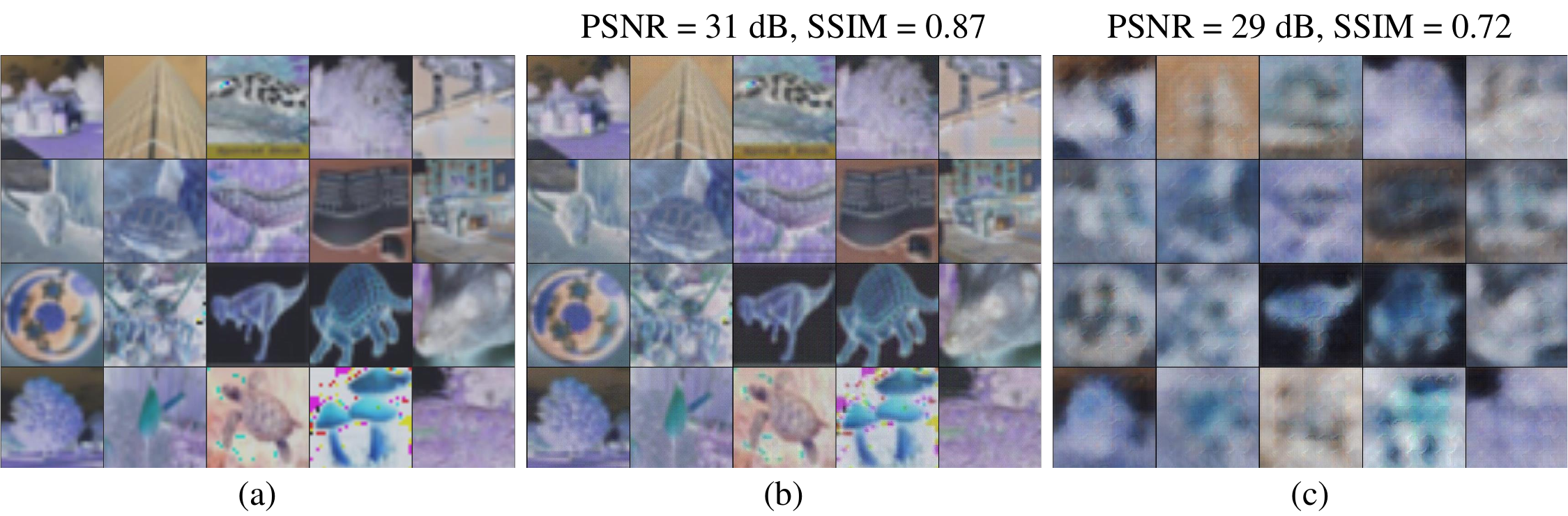}
	\caption{Image transmission results on the CIFAR-100 dataset. (a) Original images. (b) Reconstructed images based on the GSC model. (c) Reconstructed images based on the CSC model.}
	\label{fig:GSC_exp4}
\end{figure}

Fig. \ref{fig:GSC_exp1} presents the evaluation results on the MNIST dataset, where both Fig. \ref{fig:GSC_exp1}(b) and Fig. \ref{fig:GSC_exp1}(c) reconstruct the content of Fig. \ref{fig:GSC_exp1}(a). However, Fig. \ref{fig:GSC_exp1}(b) provides superior image details. The values shown above Fig. \ref{fig:GSC_exp1}(b) and \ref{fig:GSC_exp1}(c) indicate the differences in PSNR and SSIM when compared to Fig. \ref{fig:GSC_exp1}(a), with Fig. \ref{fig:GSC_exp1}(b) attaining higher scores.
Fig. \ref{fig:GSC_exp2} shows the evaluation results on the Fashion-MNIST dataset, where both Fig. \ref{fig:GSC_exp2}(b) and Fig. \ref{fig:GSC_exp2}(c) reconstruct the content of Fig. \ref{fig:GSC_exp2}(a). However, Fig. \ref{fig:GSC_exp2}(b) is noticeably clearer than Fig. \ref{fig:GSC_exp2}(c). The PSNR and SSIM results further confirm the higher quality of Fig. \ref{fig:GSC_exp2}(b).
Fig. \ref{fig:GSC_exp3} and \ref{fig:GSC_exp4} illustrate the evaluation results on the CIFAR-10 and CIFAR-100 datasets. In both cases, Fig. \ref{fig:GSC_exp3}(b) and \ref{fig:GSC_exp4}(b) accurately reconstruct the original images, while the reconstructions in Fig. \ref{fig:GSC_exp3}(c) and \ref{fig:GSC_exp4}(c) appear blurry. The PSNR and SSIM results also indicate that the images generated by the GSC models are of higher quality than those produced by the CSC models.

The superior performance of the GSC model is attributed to the advantages of the GAI model architecture (i.e., MAE), which extracts more precise semantic information compared to CNN architectures. Furthermore, due to its robust generative capability, the proposed GSC model achieves more accurate image reconstruction in SC.

\bibliographystyle{ieeetran}
\newpage